\documentclass[12pt]{article}
\usepackage[sorting=none,url=true,firstinits=true,backend=biber]{biblatex}
\pdfoutput=1
\usepackage{hyperref}
\usepackage{xcolor}
\usepackage{pdfpages}
\usepackage{pax}
\usepackage{enumitem}
\usepackage{varwidth}
\usepackage{tasks}
\usepackage{csvsimple}
\usepackage{longtable}
\usepackage[
  top=2cm,
  bottom=2cm,
  left=3cm,
  right=2cm,
  headheight=26pt, % as per the warning by fancyhdr
  includehead,includefoot,
  heightrounded, % to avoid spurious underfull messages
]{geometry} 
\usepackage{graphicx}
\usepackage{amsmath}
\usepackage[document]{ragged2e}
\usepackage{fancyhdr}
\usepackage{xcolor}
\def\eg{\emph{e.g., }}
\def\ie{\emph{i.e., }}
\usepackage{hyperref}

\fancypagestyle{firstpage}{%
  \fancyhf{}% clear default for head and foot
  \lhead{\includegraphics[width = 4cm]{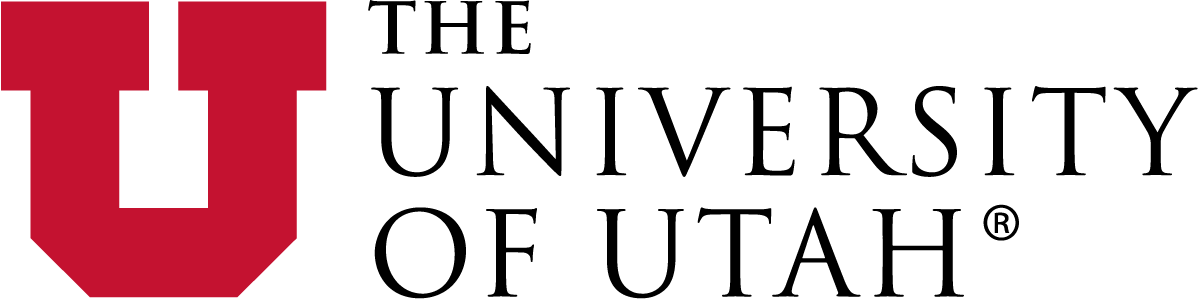}}
  \rhead{\includegraphics[width = 4.5cm]{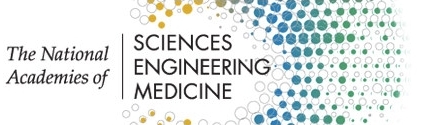}}
}
\usepackage{amsfonts}

\usepackage{titling}

\title{\textbf{Machine Learning in Heterogeneous Porous Materials}}
\author{AmeriMech Symposium Series}
\date{January 25, 2022 \\ \mbox{}\vspace{20px} \\
\noindent\makebox[\textwidth]{\protect\includegraphics[width=\linewidth]{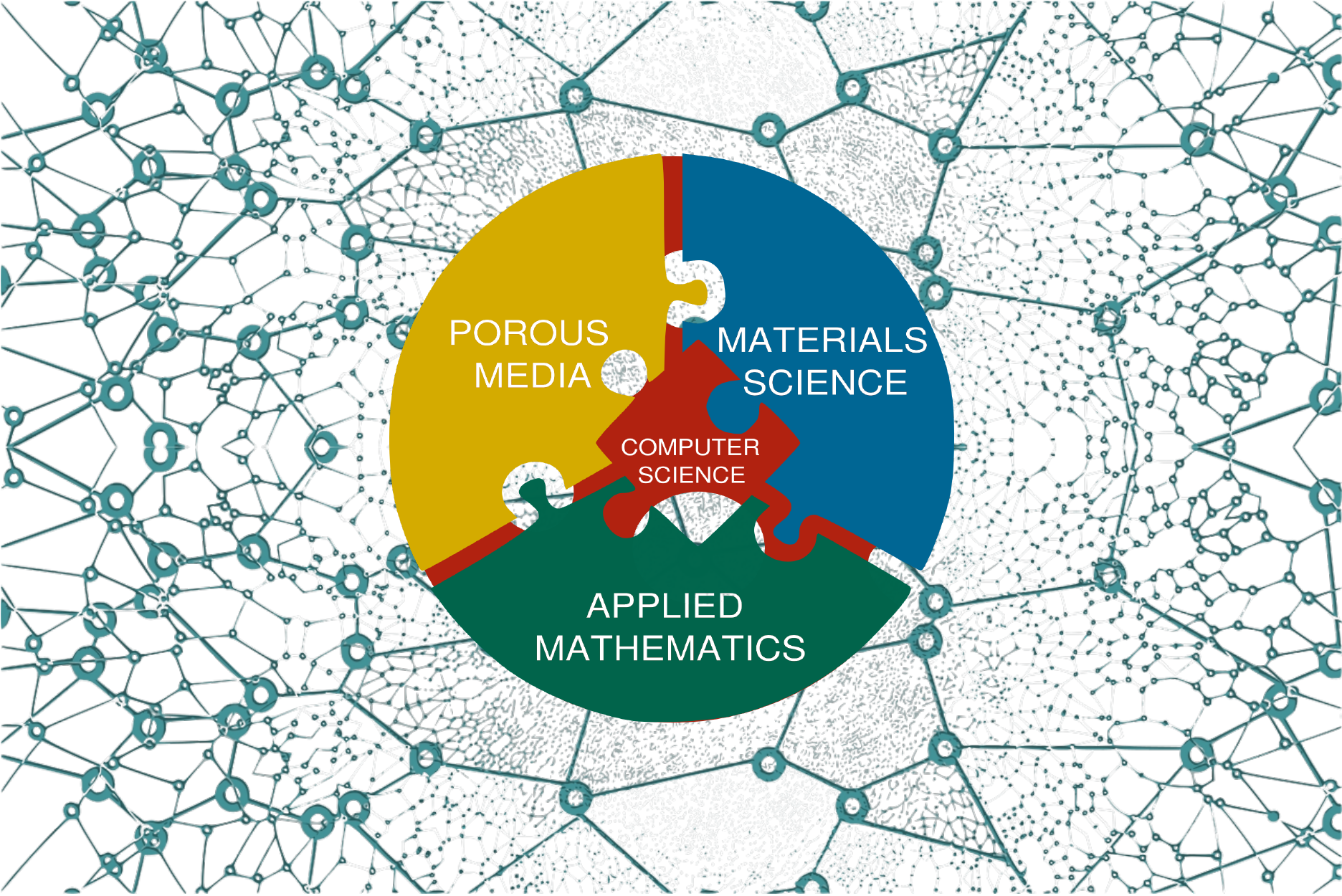}}}
\begin{document}

%Title Page
\maketitle
\thispagestyle{firstpage}
\newpage
\thispagestyle{empty}
\vspace*{\fill}
\begin{center}
 \mbox{\textit{\large This page is intentionally left blank}}   
\end{center}
\vspace{\fill}
\newpage
\thispagestyle{empty}
\tableofcontents
\thispagestyle{empty}
\newpage
\setcounter{page}{1}

%Introduction
\begin{refsection}
\section*{Introduction}
\addcontentsline{toc}{section}{Introduction}
The ``\textbf{Workshop on Machine learning in heterogeneous porous materials}" was virtually held on October 4-6, 2020 at the University of Utah. This workshop was part of the AmeriMech Symposium series sponsored by the National Academies of Sciences, Engineering and Medicine and the U.S. National Committee on Theoretical and Applied Mechanics.

The workshop for the first time brought together senior- and early- career international experts in the areas of heterogeneous materials, machine learning (ML) and applied mathematics to identify how machine learning can advance materials research.

There is no debate that machine learning has tackled many scientific and engineering problems in the past decade. The explosion of machine learning methods developed by industry (e.g. Google, Facebook, etc.) has been particularly successful in addressing data-rich problems such as visualization (e.g. facial recognition) where machine learning techniques are very effective due to their ability to interpolate and fit using big data for training. However, their role in multi-physics, multi-scale problems, which are often data sparse and require extrapolation (e.g. prediction, forward modeling), is less clear. For prediction and forward modeling, the underlying governing equations are often critical. Therefore, physics-informed machine learning approaches that combine the underlying equations and physical constraints with data-driven approaches are needed for many scientific problems. These approaches are far less developed by industry and therefore present a key knowledge gap that academia and national labs can fill. In addition, scientific discoveries, energy resiliency, and national security require an in-depth understanding of multi-scale, multi-physics and heterogeneous processes in order to predict and eventually control system behavior. 

\begin{figure}[ht!]
    \centering
    \includegraphics[scale = 0.5]{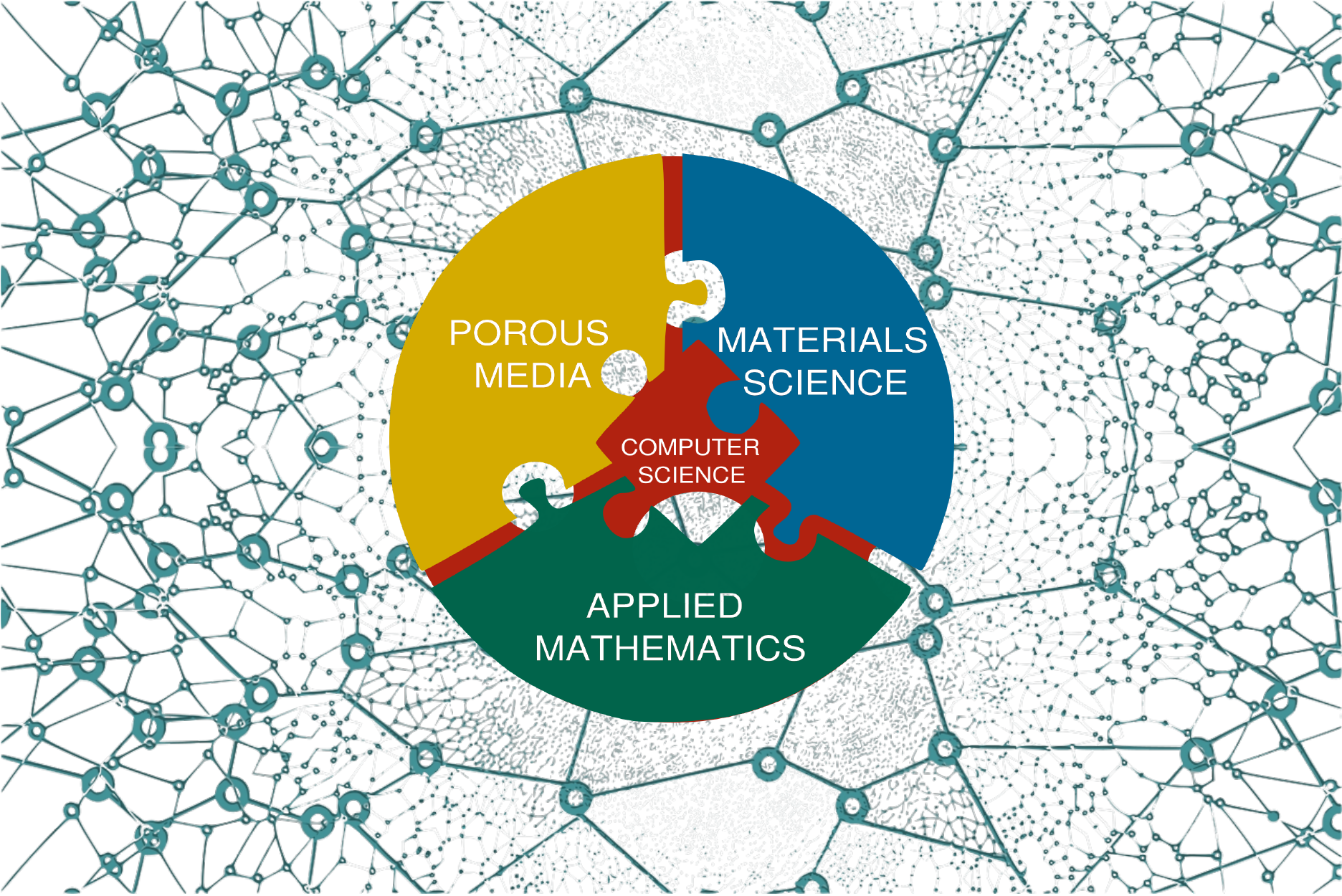}
    \caption{ML unites three scientific communities.}
    \label{periodicBC}
\end{figure}

The workshop's goal was to bring three scientific communities of applied mathematics, porous media, and material sciences together to:

\begin{enumerate}
    \item Discuss the state-of-the-art in each community,
    \item Promote crosstalk and accelerate multi-disciplinary collaborative research,
    \item Identify challenges and opportunities.
\end{enumerate}

The workshop identified four topic areas (\textbf{TAs}) that would benefit from machine learning enabled cross-disciplinary research between all three communities. These TAs are:

\begin{itemize}
    \item \textbf{TA 1:} ML in predicting materials properties, and discovery and design of novel materials
    \item \textbf{TA 2:} ML in porous and fractured media and time-dependent phenomena 
    \item \textbf{TA 3:} Multi-scale modeling in heterogeneous porous materials via ML
    \item \textbf{TA 4:} Discovery of materials constitutive laws and new governing equations
\end{itemize}

The workshop was attended by 70 participants in total. It featured six keynote lectures and four parallel sessions. The keynote lectures were:

\begin{itemize}
    \item ``Combining Graph Theory and Machine Learning to Characterize Fractured Systems" by Dr. Gowri Srinivasan from Los Alamos National Laboratory.
    \item ``Data-Driven Learning of Nonlocal models: Bridging Scales with Nonlocality" by Dr. Marta D'Elia from Sandia National Laboratories.
    \item ``Applications of Machine Learning Techniques in Fracture Mechanics" by Prof. Huajian Gao from Nanyang Technological University.
    \item ``Bioinspired AI towards Modeling, Design and Manufacturing of de novo Materials" by Prof. Markus J. Buehler from MIT.
    \item ``Generative Design and Additive Manufacturing of Three-Dimensional architected metamaterials" by Dr. Grace X. Gu from University of California, Berkeley.
    \item ``Learning Solution Operators In Continuum Mechanics" by Prof. Andre Stuart from Caltech.
\end{itemize}

The abstracts and biographical information of the speakers are available in Appendix A.

Each morning and afternoon session started with a keynote speech, followed by a break and discussion session. At the end of the day, a briefing session was held where all the chairs reported the highlight of their discussion to all the participants.\\

During the seven discussion sessions (three in Day 1, three in Day 2 and one in Day 3), following questions were discussed to gain consensus the state-of-the-art, challenges, and future directions:

\begin{enumerate}
    \item What is the state-of-the-art in your topic area?
    \item What are the existing challenges and opportunities in your topic area?
    \item How inclusive is the community associated with this topic area?
    \item Where would we like this area to be in ten years?
    \item List some of the common science questions in this area that overlaps with other communities (Applied Math, Porous material, Materials Science).
    \item How can your community learn from others to overcome some of your existing challenges?
    \item What technical advances must be made to overcome the existing challenges?
    \item How can we make our communities more inclusive?
\end{enumerate}

The symposium was also focused in promoting JEDI (Justice, Equity, Diversity, and Inclusion). To this end, a diverse group of keynote speakers, chairs, and participants from different disciplines and at different stage of their careers were invited to this symposium. Furthermore, there was some discussion within each sessions with JEDI focus.\\

The following chapters summarize the discussions within each TA and provide a roadmap for the future research directions within their area of research.

\textbf{\href{https://newell.mech.utah.edu}{Pania Newell}}
The University of Utah\\
\textbf{\href{https://www.brown.edu/research/projects/crunch/george-karniadakis}{George Karniadakis}}
Brown University\\\textbf{\href{https://www.lanl.gov/search-capabilities/profiles/hari-viswanathan.shtml}{Hari Viswanathan}}
Los Alamos National Laboratory

%references
\printbibliography
\newpage
\end{refsection}

%TA1 - Discovering new governing equations using ML
\begin{refsection}
\begin{center}
\section*{Discovering new governing equations using ML}
\addcontentsline{toc}{section}{Discovering new governing equations using ML}
\author*{M. D'Elia$^1$, A. Howard$^2$, R. M. Kirby$^3$, N. Kutz$^4$, A. Tartakovsky$^5$, H. Viswanathan$^6$} 
\end{center}

\noindent
$^1$ Sandia National Laboratories, CA, USA.\\
$^2$ Pacific Northwest National Laboratory, WA, USA.\\
$^3$ The University of Utah, UT, USA.\\
$^4$ University of Washington, WA, USA.\\
$^5$ University of Illinois, Urbana-Champaign, IL, USA.\\
$^6$ Los Alamos National Laboratory, NM, USA.

\subsection*{Abstract}
\addcontentsline{toc}{subsection}{Abstract}
A hallmark of the scientific process since the time of Newton has been the derivation of mathematical equations meant to capture relationships between {\em observables}. As the field of mathematical modeling evolved, practitioners specifically emphasized mathematical formulations that were predictive, generalizable, and interpretable. Machine learning's ability to interrogate complex processes is particularly useful for the analysis of highly heterogeneous, anisotropic materials where idealized descriptions often fail. As we move into this new era, we anticipate the need to leverage machine learning to aid scientists in extracting meaningful, but yet sometimes elusive, relationships between observed quantities.   

\subsection*{Introduction}
\addcontentsline{toc}{subsection}{Introduction}
The derivation of governing equations for physical systems has been dominating the the physical and engineering sciences for the past decades. These governing equations are used for predictive physics-based models to forecast system behavior. Indeed, this is the dominant paradigm for the modeling and characterization of physical processes, engendering rapid and diverse technological developments in every application area of engineering and the sciences. Until the middle of the $20^{\rm th}$ century, many models of natural and engineered systems relied on linear governing equations that are amenable to analytic solutions. With the invention of the computer and the rise of scientific computing, nonlinear problems could easily be explored through numerical simulation using techniques such as finite elements. Scientific computing allows one to emulate diverse and complex systems that are high-dimensional, multiscale, and potentially stochastic in nature. In modern times, the rapid evolution of sensor technologies and data-acquisition software/hardware, broadly defined, has opened new fields of exploration where governing equations are difficult to generate and/or produce. Biology and neuroscience, for instance, easily come to mind as application areas where first-principals derivations are difficult to achieve, yet data is now becoming abundant and of exceptional quality. The coarse-grained macroscopic behavior of heterogeneous materials is also often difficult to derive or characterize from known microscopic descriptions. The ability to discover governing equations directly from data is thus of paramount importance in many modern scientific and engineering settings as such equations provide both interpretability and generalizability.  

Here we first summarize the state of the art in the area of machine learning (ML) dedicated to deducing governing equations from data, highlighting the importance of three fundamental features of such formulations (which we will define below): predictivity, generalizability, and interpretability.
We then explore the current knowledge gaps and challenges to making this area fully realizable, and we provide some recommendations on where the field might focus its efforts to enable rapid advancements.
\subsection*{State-of-the-art}
\addcontentsline{toc}{subsection}{State of the art}
In this section we review recently developed mathematical architectures that discover governing equations directly from data. In this context, in a short time, the diversity of methods and their capabilities have had an impact across a number of application areas.
In all cases, measurements for which the signal-to-noise ratio is low (e.g. noisy data) represent significant challenges for any analysis of the underlying signal, including model discovery. Two machine learning areas have dominated the discovery of parsimonious governing equations:  symbolic regression (SR) with genetic programming (GPSR) and sparse regression. These two areas draw from the broader class of dictionary learning methods and have been employed in a wide range of applications outside the materials science field, as described below.  We anticipate that both of these methods (as well as future methods that evolve from these) can be adapted in the future to heterogeneous anisotropic multiscale problems as found in materials science.

An early method for the discovery of governing equations was symbolic regression~\cite{Bongard2007pnas,Schmidt2009science}. SR is a method that aims to model an input dataset without assuming its form. Instead, candidate models are proposed and evaluated by the algorithm, and the only assumption is that the data can be modeled by some algebraic expression. This is in contrast to traditional regression methods in which model form selection is made first and the regression method then estimates the model parameters. From a general perspective, SR is an optimization problem that occurs over a non-numeric domain of mathematical operators (\eg $+$, $\exp$, $\sin$, $d/dx$, 
{\em etc.}) and numeric domain of model parameters. The SR model is characterized by a variable-length combination of operators and parameters and, therefore, poses an infinite space of possible model forms to search. In practice, the mathematical operator domain is limited by a finite set of operations and a limited model length (\ie stack size). Note that SR is computationally expensive, especially when used with genetic programming to search for models. 
SR can also be used in a recursive multidimensional setting with physics-informed neural networks (described below) to discover governing relationships among the key variables of a given problem~\cite{udrescu2020ai}. Like SR for dynamics, it is expensive computationally, but allows for the discovery of many physics relations and equations, e.g. 100 equations from Feynman’s Lectures on Physics.

Genetic programming is the most commonly used model-evolution algorithm for SR, termed GPSR. Genetic algorithms (GA) are incredibly flexible and powerful optimization schemes \cite{Koza_1994} that attempt to mimic the evolutionary selection processes observed in natural systems. Genetic programs are a type of GA in which models are represented as (nested) variable-length tree structures representing a program instead of a fixed-length list of operators and values. Within GPSR, genetic programs are used to generate random perturbations to models which are evaluated against a fitness function(s). This fitness is used to select models most likely to perform better, and then randomly recombine (\ie crossover) and permute (\ie mutate) them to generate new candidate models. At the same time, the candidates with the poorest fitness are evolved out of the population (\eg natural selection). The iterative exploration of the solution space is subject to both randomization and guidance from the particular fitness, crossover, and mutation procedures implemented. In applying GPSR, it is important to keep in mind that this technique is not scalable for large-scale problems, which may limit its applicability to future engineering and scientific applications.

As an example of the application of GPSR to multiscale systems, this method was shown to learn the symbolic expression of the von Mises yield surface, provided corresponding training data from simulations of representative volume elements~\cite{Bomarito21}. Extension to learning the evolution equation for state variables (\eg plastic strain) was also demonstrated. While this demonstration ``learned'' an existing model, it demonstrated that GPSR can satisfy an engineering requirement: a means for verification of artificial intelligence (AI) and ML models and assessment for meaning and insight.  More recently, researchers have used GPSR to learn microstructure-dependent plasticity models for additively-manufactured Inconel 718 \cite{Garbrecht21}. These learned models were automatically parsed within a topology optimization code; a capability that can be included within various software environments.

Following on from the data-regression approaches mentioned above, the {\em sparse identification of nonlinear dynamics} (SINDy) method~\cite{Brunton2016pnas} is a SR method that leverages time-series data to discover the governing equations from a library of candidate models.  The sparse regression procedure extracts the terms which best represent the time-series data. The SINDy algorithm has been broadly applied to a wide range of systems, including for reduced-order models of fluid dynamics~\cite{Loiseau2017jfm,Loiseau2018jfm,loiseau2020data,guan2020sparse,deng2021galerkin,callaham2021empirical,callaham2021role} and plasma dynamics~\cite{Dam2017pf,kaptanoglu2020physics}, turbulence closures~\cite{beetham2020formulating,beetham2021sparse,schmelzer2020discovery}, nonlinear optics~\cite{Sorokina2016oe}, numerical integration schemes~\cite{Thaler2019jcp}, discrepancy modeling~\cite{Kaheman2019cdc,de2019discovery}, boundary value problems~\cite{shea2021sindy}, identifying dynamics on Poincare maps~\cite{bramburger2020poincare,bramburger2021data}, tensor formulations~\cite{Gelss2019mindy}, and systems with stochastic dynamics~\cite{boninsegna2018sparse,callaham2021nonlinear}. In the work on plasmas~\cite{Dam2017pf}, for instance, a predator-prey type dynamical system that approximates the underlying dynamics of the three energy state variables was discovered. Importantly, the model is amenable to a bifurcation analysis that reveals consistency between the bifurcation structures observed in the  data.
The integral formulation of SINDy~\cite{Schaeffer2017pre,messenger2021weak} has also proven to be powerful, enabling the identification of governing equations in a weak form without recourse to computing derivatives; this approach has recently been used to discover a hierarchy of fluid and plasma models~\cite{Reinbold2020pre,gurevich2019robust,alves2020data,reinbold2021robust}. 
The open source software package, PySINDy\footnote{\url{https://github.com/dynamicslab/pysindy}.}, has been developed in Python to integrate the various extensions of SINDy~\cite{deSilva2020JOSS}. An attractive feature of SINDy is that it simply solves an over-determined ${\bf A}{\bf x}={\bf b}$ by promoting sparsity and making it modular and amenable to enabling innovations. Moreover, it is exceptionally efficient computationally in comparison with SR, thus allowing for discovery on data with orders of magnitude less computational time. It can also be used with neural network (NN) architectures which provide automatic differentiation~\cite{rackauckas2020universal,gelbrecht2021neural} and learning coordinates and models jointly~\cite{Champion2019pnas,kalia2021learning}. 

In addition to the symbolic (dictionary) methods mentioned above, there is a strong surge of interest in physics-informed machine learning (PiML) \cite{raissi2021physics} by means of deep neural networks (DNNs). This area evolved concurrently with the aforementioned dictionary learning methods. Interest in DNNs within the scientific community started in part due to their flexibility and expressiveness (see \cite{rico1994continuous}, for a first work in this direction, and \cite{raissi2018multistep}). DNNs were first used as physics-constrained regressors such as physics-informed NNs (PINNs); in this context, we also mention their combination with reduced order modeling \cite{kim2021fast}. However, more recently, their use has morphed into the topic of operator learning.  Given this trajectory, some anticipate that this line of work will transform what we as a community mean by ``learning governing equations." In this context, a recent review article \cite{KarniadakisPINNs2021} summarizes the state of the art of PINNs and DeepONets \cite{lu2021learning}, that we describe below. In PINNs-type approaches 
\cite{raissi2019physics,raissi2017physicsI, raissi2017physicsII}, the solution of a {\it known} PDE is modeled by a DNN whose parameters, together with other model parameters, are to be learned. We stress that, for these approaches, only constitutive relationships and model parameters are ``discovered'', whereas the fundamental underlying physics is established {\it a priori} (e.g., conservation laws). More recently, building on PINNs and other learning paradigms, new deep learning tools designed to discover governing relationships, have been considered. Among these we mention {\it neural-operator}-type approaches \cite{li2020fourier,li2020multipole,li2020neural} and DeepONet \cite{lu2021learning}. This type of deep learning broadens the concept of ``governing equations" to include complex mathematical forms such as DNNs, and, in doing so, challenges traditional notions of interpretability, as addressed below.

\begin{figure}[h!]
    \centering
    \includegraphics[width=\textwidth]{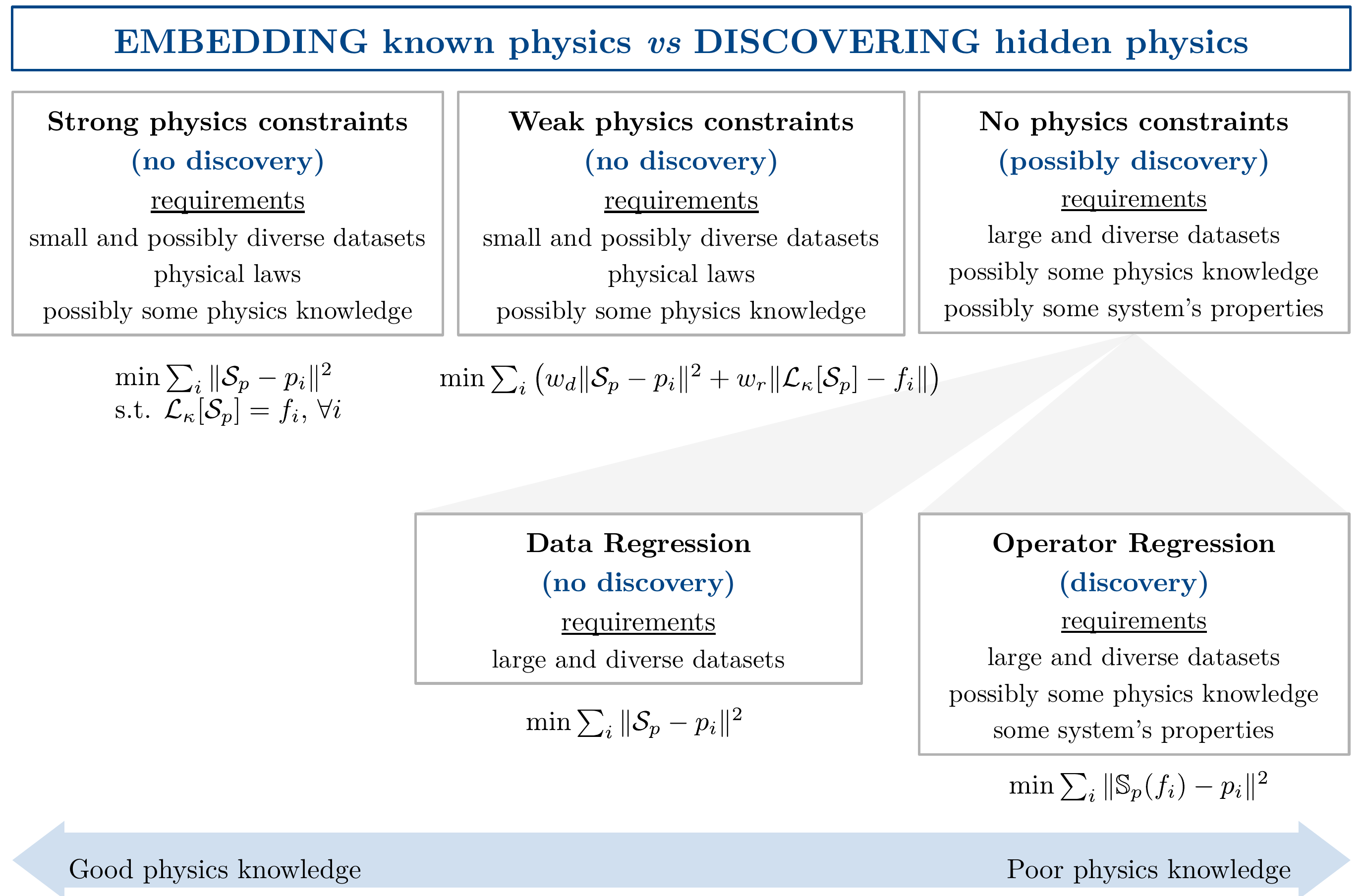}
    \caption{Different approaches to learn surrogates from regression to discovering new physics. The assumption is that (input,output) observation pairs $(f_i,p_i)$ are available; $f$ is an external source and $p$ is the quantity of interest of a potentially unknown physical system. $\mathcal S_p$ is a pointwise surrogate for the quantity $p$ whereas $\mathbb S(f)$ is a surrogate for the solution map that given the input $f$, returns a solution $p$. In the top-left box, the known physics, given by the residual equation $\mathcal L p-f=0$, is embedded strongly for each sample via equality constraints. In the top-center box, the known physics is embedded weakly as the penalization of the residual equation in the loss function, weighted by the parameter $w_r$. In the top-right corner, no prior knowledge on the physics is assumed; two possible approaches are reported at the bottom. In the bottom-left box we have simple data regression that returns a pointwise surrogate for $p$. In the bottom-right box we have operator regression that returns an input-output map that acts on function spaces and for every input $f$ returns an output $p$. This box is the only one that enables the discovery of hidden physics.}
    \label{fig:physics}
\end{figure}

We summarize possible ways to embed or discover physics via machine learning in Figure \ref{fig:physics}. 
Here, several approaches are listed for decreasing levels of confidence in the knowledge of the underlying physics and, for each of them, we indicate where physics discovery is possible and what type of datasets are required. In the top-left box, {\it known} physics laws are embedded in the learning process via ``strong'' constraints, \ie via equality constraints to the optimization problem. In the top-middle box, physics constraints are embedded ``weakly'', as part of the cost functional (or loss functional) to be minimized. This approach is typical in PINNs-type algorithms \cite{He2020AWR,pang2020npinns,pang2019fpinns,Reyes2021PRF,Tartakovsky2020WRR,Tartakovsky2020JCP_PICKLE,Xu2021CMAME} and in {\it nonlocal-kernel-regression} algorithms \cite{Xu2021,You2021,you2021MD,You2020Regression}. Both approaches rely on some underlying physical knowledge in the form of an equation or a constitutive law. The top-right box represents those approaches for which physics knowledge is poor or absent. In this case, learning solely relies on data and, for this reason, it requires even richer datasets. Note that in this category we have two types of learning (represented by the two bottom boxes): a pure data-regression approach, where given some dataset, a surrogate (\eg a NN) is trained in a least square sense, and an input-output approach, where, given input-output pairs for a specific system, the surrogate ``discovers'' new constitutive input-output relationships. In the latter class, we have both symbolic (dictionary) methods such as SR and SINDy and operator learning approaches such as neural operators and DeepONets.

\paragraph{\underline{Prediction, Generalization and Interpretability}}

As the use of AI/ML technologies grows, there is an increasing number of people who believe that AI/ML representations will surpass model-based prediction capabilities.  This enthusiasm has been bolstered by computer science success stories around DeepMind's AlphaGo and Tesla's self-driving car technologies. Critics of this view normally point to three desired features of models; they desire models that are predictive, generalizable, and interpretable. In addition, for these success stories, large training data sets exist whereas they often do not for heterogeneous materials problems. Since there is still debate about the precise definitions of these terms (and the debate is sometimes a consequence of the field or subfield of science in which they are used), we purposely give here conceptual definitions that inform our discussion below and that are not intended to be sharp. A model is considered {\em predictive} if it can forecast likely future outcomes.
A model is considered {\em generalizable} if the model can adapt to new, previously unseen data (\ie data outside the training set); this property can also be referred to as the ability to extrapolate.\footnote{We acknowledge that there is overlap between predictive power and generalization, but leave them as separate but complementary concepts.}
A model is considered {\em interpretable} if relevant 
(to the problem at hand) understanding of the relationships either contained in data or learned by the model can be inferred \cite{Murdoch22071}.  

Some AI/ML advocates in the computer science community argue that the hallmark of science is not ``understanding" {\em per se} in a reductionist sense, but rather ``trust" in generalization and predictive capability. When we are able to more faithfully generalize AI/ML results, might their use with materials science systems become more ubiquitous? With more generalizable AI/ML methods, difficult problems such as multi-scale, heterogeneous problems that lack traditional governing equations may be within reach. We acknowledge that at present, there is a tension between these two camps:  those that stress predictive power while allowing a sacrifice of interpretability, and those that emphasize interpretability over predictive power. Both camps appear to agree on generalizability; however, they disagree as to whether predictive power or interpretability most naturally leads to capturing this constraint.

\subsection*{Knowledge gaps and challenges}
\addcontentsline{toc}{subsection}{Knowledge gaps and challenges}
Despite the recent advances in Scientific Machine Learning (SciML),
and particularly the use of ML techniques such as dictionary learning, to learn governing equations, several modeling and computational challenges hinder their general usability in practical contexts, including the simulation of heterogeneous porous materials. We list and discuss seven challenges, some of which lead directly to recommendations for action. We have highlighted these directly in this section; several recommended areas for action are further discussed in the following section.

The first two challenges are 1) the {\it interpretability} of the machine-learned surrogates and 2) their {\it generalization} to settings (boundary conditions, environment conditions such as temperatures, body loadings, etc.) that are substantially different from the ones used during training. These are not necessarily related to the application of interest in this report but are common gaps in the discovery of new governing equations for several engineering and scientific applications.
Other challenges, strongly related to the simulation of materials, are 3) addressing the {\it multiscale} nature of the physical systems of interest and of the available data, 4) dealing with the presence of high degrees of {\it heterogeneity} at different scales, and 5) incorporating model or data {\it uncertainty} in the learning algorithms. We also briefly report on an additional challenge that is not directly related to the ML strategy, but highly affect its outcome, \ie 6) the need for efficient optimization techniques.

\begin{figure}[h!]
    \centering
    \includegraphics[width=\textwidth]{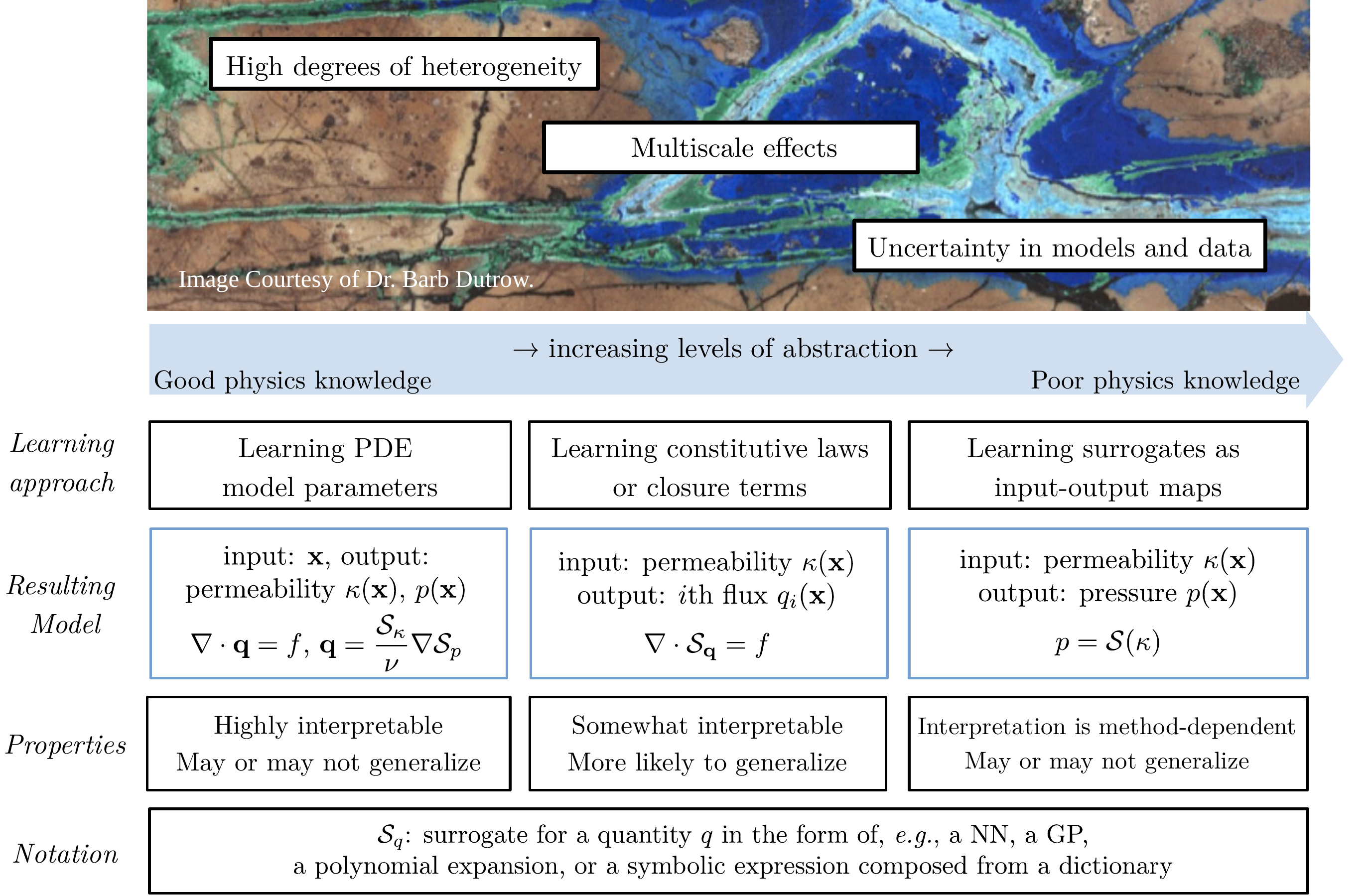}
    \caption{For subsurface flow, we list three possible learning approaches, their corresponding surrogates, and their properties. ML techniques are listed for increasing levels of abstraction and physics knowledge.}
    \label{fig:learning-gov-eq}
\end{figure}

A summary of the discussion that follows can be found, for the case of subsurface transport through porous media, in Figure \ref{fig:learning-gov-eq}. This specific application was chosen as an outstanding representative of all of scenarios considered in this report. The complexity of the subsurface environment and the difficulty in accessing such an environment, are such that dealing with multiscale effects, heterogeneity of the medium, and uncertainty in the measurements and in the models is a nontrivial task. In this figure, we list three possible learning approaches for increasing levels of abstraction of the resulting surrogate, $\mathcal S$, and we highlight their properties in terms of interpretability and generalization capability. Specifically, $\mathcal S_q$ represents a surrogate for a quantity $q$ in the form of, \eg a NN, a Gaussian Process (GP), a polynomial expansion, or a symbolic expression composed from a dictionary. The models listed here are by no means a fully representative set of state-of-the-art methods, but provide valuable examples of possible approaches characterized by different degrees of interpretability and generalizability. In the first column, the solutions of {\it known} PDEs, such as Darcy and advection-dispersion equations, are modeled as surrogates together with other model parameter fields (e.g., permeability). Representatives of this technique are PINNs-type approaches \cite{He2020AWR,Tartakovsky2020WRR}.  Clearly, the resulting surrogate is a solution of a PDE whose terms have an understood physical interpretation. On the opposite side of the spectrum of physics knowledge, we have the approach illustrated in the last column where the resulting model is an operator (or map) from some of the system's inputs to the solution, the pressure in this case. Among this class of techniques, we mention neural-operator-type approaches as well as symbolic (dictionary) methods. In this case, the interpretability depends on the method: as explained above, neural-operator-type approaches are harder to interpret whereas the interpretability of the components of a symbolic method is more straightforward.

\paragraph{\underline{Interpretability}}
By interpretability, many scientists mean the ability of identifying in the resulting surrogate model a physical behavior such as diffusion, advection, reaction, etc. While this is particularly straightforward when learning coefficients of a known PDE, it is much less intuitive when learning, \eg NNs, as surrogates of a solution operator. In fact, while equations are highly interpretable, the architecture of a NN is not directly connected to a physical phenomenon. When little is known about the physics of the system at hand, the lack of interpretability affects the extent of trust in the ML surrogate.

The idea of interpretability stated above is not incorrect, but incomplete. We hold that the lens through which one should view interpretability is {\em decision-making}. For purposes of example, consider the following. The actions supporting decision-making often involve directly or implicitly enumerating all possible outcomes of a particular choice. Prior to the modern age, this enumeration step often involved speculation and extrapolation built upon a combination of historical pattern analysis and human experience. The decision-maker's level of trust in the predictions was in part determined by their trust in the person from whom they were receiving their information and upon the {\em resonance} or {\em dissonance} of those predictions with the decision-maker's own experiences and intuitions. In the modern age, both simulation science and data science have augmented the outcome prediction step. Although the use of mathematical and statistical predictors based upon a combination of first-principle modeling and data-driven science is now ubiquitous and has often replaced human intuition as the generator of possible future outcomes, how the decision-maker engages with these tools has not changed -- it is predicated on {\em trust}. As AI/ML tools move to the forefront of options used by decision-makers to create data-driven outcome predictions, it is important to develop strategies and tools that enable {\em explainable} AI/ML: tools that encourage and support trust-building.

There are a variety of ways by which decision-makers build trust in the tools that they use. In the case of simulation science, the most common trust enabler is an agreement on what are {\em first principles} and on how those first principle components are assembled to yield a prediction. Many decision-makers struggling with the use of AI/ML in outcome prediction either explicitly or implicitly assume that this strategy is the only strategy for interpretability. However, this bottom-up approach to interpreting predictions is not the sole standard for interpretability.  There are many systems for which, although we might vaguely understand the various first-principle building blocks that are being used, the complexity of the assembling of the system makes it impossible for us to create a cause-and-effect chain in our minds that justifies our trust. As an analogy, though one might claim that an understanding of Newton's laws of motion combined with an understanding of thermodynamics aids someone in appreciating how an automobile transmission works, it would be an exaggeration to say that a person has built trust in the performance of their transmission due to this knowledge. Similarly, a knowledge of the physics underlying a system may not be sufficient to create trust in the model.

An alternative way by which we gain experience and trust in complex systems is through {\em interrogation}. Through an understanding of the purposes of the model and the assumptions upon which it was built, we build our trust in its predictions by presenting it with scenarios for which we believe we understand what the prediction should be. We build confidence in the model each time the model, under known circumstances, reacts as we anticipate. Once this base level of trust is established, we begin to interactively interrogate the model with situations for which we may or may not have a complete idea as to what the prediction might be; however, we want to see and reason about the outcomes provided by our AI/ML recommender system. Like in the social context, this interactive session allows the decision-maker to build trust in their tools while at the same time building an appreciation of its biases and limitations. 

{\bf Recommendation:} We believe that research into interactive visualization and interrogation tools may play an important role in enabling explainable AI/ML for material systems. For example, the authors in \cite{Xing2020} introduced the use of AI/ML tools for uncovering interpretable shared ``hidden'' structures across data spaces for design space analysis and exploration. Their work demonstrated how AI/ML tools could be used within an interactive framework (dSpaceX) to first build trust in the tools themselves, and then later to interrogate new topological optimization designs. Similarly, the authors in \cite{Liu2019} introduced {\em NLIZE} -- a perturbation-driven visual interrogation tool for analyzing and interpreting natural language inference models. They introduced a visualization system that, through a tight yet flexible integration between visualization elements and the underlying model, allows a user to interrogate the model by perturbing the input, internal state, and prediction while observing changes in other parts of the pipeline. They used the natural language inference problem as an example to illustrate how a perturbation-driven paradigm can help domain experts assess the potential limitation of a model, probe its inner states, and interpret and form hypotheses about fundamental model mechanisms such as attention. 

\paragraph{\underline{Generalization}}
ML algorithms are particularly effective in ``interpolation'' tasks, \ie in generating surrogates that well-represent the dataset used during training. However, simple interpolation that only predicts the regime captured by available data is not sufficient for reliable predictions. A predictive surrogate must extrapolate (or generalize) to regimes that are different from the ones used for training. Ideally, a reliable surrogate would only embed the material's constitutive behavior and be independent of the system's inputs such as environment conditions, boundary and initial conditions, loadings, {\em etc}. Even more desirable would be a surrogate that generalizes beyond a specific material (\ie a surrogate for subsurface transport that provides reliable predictions regardless of the composition of the subsurface, as long as the system's inputs are available). Several SciML algorithms learn surrogates for the state of the system rather than for the constitutive behavior; this approach may lead to surrogates that are tied to specific inputs such as boundary conditions or that only represent solutions that belong to the training set. Recent works focused on learning constitutive laws or, more in general, a surrogate for the solution operator itself, are more likely to be independent of the system's inputs and, hence, to generalize better. 

Finally, one has to keep in mind that with higher levels of generalization and abstraction, the price to pay might be interpretability. This is the case of {\em e.g.}, {\it neural operators}, \ie NNs that reproduce the system's behavior, where the surrogate is a NN itself and little can be said about its connection to a physical phenomenon, as anticipated above. 

{\bf Recommendation:} A particularly complex task is the prediction of emergent phenomena; most of the current SciML algorithms are still not able to capture anomalies that may arise in a system and that are not accounted for in the training set, such as bifurcations. This indicates an area of future research.

\paragraph{\underline{Multiscale nature, heterogeneity, and stochasticity}} Discovery of equations for deterministic, homogeneous physical systems characterized by a single time and length scale has experienced significant progress by means of standard mathematical physics tools (\eg thermodynamics, statistical physics, homogenization, etc.). For these problems, we expect SciML to be successful in discovering governing equations; in fact, in the absence of multiscale effects, uncertainties (in the model and data), and heterogeneities, we have equations that model such systems and can deliver accurate, efficient, and predictive simulation tools in the area of materials science. In this context, ML is useful when some model parameters are unknown. However, in the presence of multiscale effects, heterogeneities, and uncertainties, current PDE-based models may be insufficient to appropriately capture the system's behavior. Thus, current SciML algorithms may fail to be predictive because they were designed and tested on a single-scale, homogeneous toy problem whose physical behavior is well-known and understood. 
In the presence of heterogeneity and when the small scale behavior affects the system's global behavior, there is the need for new surrogates that are able to capture the effects of the small scales at the continuum level. A typical example of this situation is subsurface modeling where high degrees of heterogeneity and small scale effects that cannot be captured at large scales compromise the ability to describe the system using classical models. In this case, standard PDE-based ML algorithms may fail to be predictive by addressing one scale at a time. In such cases, a way to circumvent this challenge is to learn a NN as a surrogate of the solution map; this approach exploits the ability of NNs to capture complex behavior thanks to their {\it compositional nonlinearity}. 
On the other hand, high degrees of heterogeneities, especially at the small scales, cannot be accurately detected due to difficulties in measuring material properties at very fine scales or in reaching specific locations (\eg deep subsurface layers). When this is the case, heterogeneity can be treated as uncertainty; as such, material properties and system's input are treated as random fields for which prior statistical information is available \cite{meng2021learning}. 

{\bf Recommendation:} There is a critical need for additional ML tools that can learn models while embedding prior stochastic information and deliver probability distributions, rather than deterministic surrogates. In this context, as for any other uncertainty-quantification algorithm, a ML learning tool might suffer from the so-called curse of dimensionality. This creates the need for new tools that can embed uncertainty while featuring a cost that scales linearly (or sub-linearly) with the number of parameters. Such tools should be able to quantify the total uncertainty including aleatoric uncertainty due to noisy data, epistemic uncertainty due to physics and ML models and make the effective use of the prior statistical information.

\paragraph{\underline{Efficient optimization algorithms}}
The latest successes in ML, in general, and deep learning, in particular, are in large part due to the recent advances in optimization algorithms, including stochastic gradient descent. However, the stochastic nature of such algorithms in combination with the non-convexity of the optimization problems in deep learning makes ML predictions uncertain even when the underlying physical system is deterministic (\eg a system with fully known governing equations and parameters). In addition, ML predictions might strongly depend on hyper-parameters; this fact further complicates their interpretability.    

\subsection*{Recommendations to advance the field in ten years} 
\addcontentsline{toc}{subsection}{Recommendations to advance the field in ten years}

We envisage a day when the field of mathematical modeling regularly uses AI/ML tools to aid in finding and/or refining new governing equations for complex, multiscale heterogeneous systems. To arrive at such a place in the future, we hold that several foundations areas must be addressed, as summarized below. 

\paragraph{\underline{Developing a theory of the rigorous mathematics behind ML}} 
Our primary recommendation is a call to develop a better understanding of the mathematical theory behind machine learning tools so we can design trustworthy computational capabilities. In particular, there are community members who stress that the success of the finite element method was not merely its applicability, but the hand-in-glove nature of theory and application that developed along its evolutionary path. There is a need to increase our confidence in ML models analogous to the theoretical understanding of FEM methods, where rigorous theory guarantees and implementations convergence. Such developments are necessary in order to provide rigorous bounds and uncertainty in modeling efforts, estimates for the total error, and approximation, generalization, and optimization algorithms. 

\paragraph{\underline{Advancing the understanding of generation and limitations of training data}} 
In comparison with many current applications where an abundance of data is available, such as computer vision problems, the problems considered here have limited data and/or measurement availability. As of now, the literature does not offer ML algorithms that are able to handle multiscale, multi-source, multi-resolution data in an efficient and automatic way. Relevant problems in the field require dataset curation for complex, time-dependent problems. Large datasets cause slower training, while small datasets may not sufficiently capture key features. Multi-fidelity or multi-modality data presents challenges in how to incorporate or weight different types and sources of data.
 
Two additional obstacles are the ability to deal with datasets that are large in size, unstructured, and non-regular and the ability to use datasets that are small, sparse, and irregular in sampling.  The former is particularly problematic when learning PDEs, as the solution is expected to feature some degree of smoothness. This indicates an area where further research investment is needed.

In data constrained modeling, knowledge of the underlying physics may be used to restrict the solution to within a given space. In all cases, data are absolutely the most important asset for leveraging the various approaches advocated.

\paragraph{\underline{Developing standardized benchmark datasets}} 
Many fields have introduced standard benchmarks to compare and build confidence in results from numerical methods. For machine learning, an early standard benchmark was the MNIST handwritten digit database \cite{lecun-mnisthandwrittendigit-2010}. More sophisticated data sets, including the ImageNet challenge \cite{deng2009imagenet}, and the Penn Machine Learning Benchmark for classification \cite{olson2017pmlb}, have provided a platform for the proper evaluation of methods. Many more benchmarking suites are currently being established for new machine learning applications, such as the Open Graph Benchmarks for machine learning on graphs \cite{hu2020open}. Benchmark problems allow comparisons between methods by providing a standard database. Currently, no such benchmark databases exist for discovering governing equations. This puts the onus on researchers to invest time and resources developing their own datasets and does not allow for standardized comparisons between methods. We suggest that there is a need for a rigorous hierarchy of challenge problems to be developed for heterogeneous, anisotropic materials. These benchmark problems should cover the range of regimes from data-rich to data-poor, include applications with and without knowledge of the underlying physics, and range from toy problems to very complex problems. 

Our overview suggests that a large toolbox of ML learning technologies is now available and ready to use to study heterogeneous porous materials and discover new governing equations. However, there is a critical need to classify and structure these tools and tightly integrate them with cleaned and curated data. Ultimately, the success of ML learning in discovering underlying equations will critically depend on building trust through transparency, data-sharing, code-sharing, and benchmarking within the community and beyond.

\section*{Acknowledgments}
\addcontentsline{toc}{subsection}{Acknowledgments}
Sandia National Laboratories is a multi-mission laboratory managed and operated by National Technology and Engineering Solutions of Sandia, LLC., a wholly owned subsidiary of Honeywell International, Inc., for the U.S. Department of Energy’s National Nuclear Security Administration under contract DE-NA0003525. This report, SAND2021-14093 R, describes objective technical results and analysis. Any subjective views or opinions that might be expressed in the paper do not necessarily represent the views of the U.S. Department of Energy or the United States Government. 

Pacific Northwest National Laboratory is operated by Battelle for the DOE under Contract DE-AC05-76RL01830.

\addcontentsline{toc}{subsection}{References}
{\renewcommand{\markboth}[2]{}% Remove header adjustment
\printbibliography}

\newpage
\end{refsection}
%TA3 - Multi-scale modeling in heterogeneous porous materials via ML
\begin{refsection}
\begin{center}
\section*{Multi-scale modeling in heterogeneous porous materials via ML}
\addcontentsline{toc}{section}{Multi-scale modeling in heterogeneous porous materials via ML}
\author*{H. Lu$^1$, C. Fraces$^1$, H. Tchelepi$^1$, D. M. Tartakovsky$^1$}
\end{center}
\noindent
$^1$Department of Energy Resources Engineering, Stanford University, Stanford, CA, USA

\subsection*{Abstract}
\addcontentsline{toc}{subsection}{Abstract}
Machine learning has been increasingly recognized as a promising tool in solving subsurface problems. There is no doubt that this technique has been incredibly successful in computer science, where large amounts of data are available and free of charge. However, machine learning faces various challenges in dealing with complex physical systems, especially when multi-scale, multi-physics phenomena present with sparse data in heterogeneous porous media. This is a field where classical physics-based simulations have been irreplaceable since the underlying governing equations are considered, which describe the phenomena of interest. We discuss the state of the art of multi-scale modeling in heterogeneous porous materials via machine learning -- the fine-scale physics can be efficiently represented by ML surrogate models and incorporated in the coarse-scale physics. Several challenges in the availability of data, incorporation of physics, and physics-informed neural network (PINN) formulations for forward problems are addressed with our recommendations to advance the field with joint forces of different communities in the future.

\subsection*{Introduction}
\addcontentsline{toc}{subsection}{Introduction}
The multi-scale and multi-physics nature is the key to understanding and managing the behavior of natural and artificial heterogeneous porous materials. This behavior dominates multiple phenomena of practical significance, such as subsurface storage of radioactive waste or carbon dioxide, sustainable exploitation of groundwater resources, and design of novel materials for electrical storage or desalination membranes. Most, if not all, of these applications rely on highly nonlinear models that strive to capture processes that occur on a multiplicity of scales both in space and time. For example, tightly coupled models used in performance assessment of high-level radioactive waste storage deals include components whose scales range from days and millimeters to hundreds of thousands of years and tens of kilometers. 

An assemblage of individual components of a multi-physics phenomenon into a complete multi-physics simulator typically ignores feedback between components or couples them loosely. The prediction errors of this strategy are well documented, especially for highly nonlinear problems in which random noise/fluctuations generated by one of the components materially affect the behavior of another~\cite{alexander-2005-algorithm, alexander-2005-noise, taverniers-2014-noise}. The full coupling of individual components ameliorates this problem but is rarely used in practice because of the high computational cost of required iterations.

In multi-scale simulations of complex processes in porous media, we are often concerned with the impact of pore-scale properties, such as pore-size distribution or pore-network topology, on large-scale material characteristics, such as hydraulic conductivity or electric capacitance. This information is required to predict biological, physical, or chemical processes in natural porous media~\cite{winter-2001-theoretical} and to design novel materials with desired macroscopic characteristics~\cite{zhang-2017-optimal}.  

\subsection*{State of the Art}
\addcontentsline{toc}{subsection}{State of the Art}
The state of the art of multi-scale modeling in heterogeneous porous materials via ML is the application to propagate the relevant information across the scales~\cite{boso-2018-information, hall-2021-ginns, taverniers-2021-mutual}. Since the ML-based surrogates and reduced-order models (ROMs) of the process-based models of individual components have often-negligible costs, their use in fully coupled multi-physics simulations might overcome the computational bottleneck~\cite{kluth-2020-deep, lu-2021-data}. ML tools, such as Bayesian nets, neural networks, and information-theoretic tools, show advantages in analyzing data structure and capturing physics features. Together with physics modeling and increased data availability, ML can be used to explore the dominant physics at different scales. It has been successfully used to solve inverse problems, e.g., inferring material properties or identifying transport sources, where traditional numerical solvers are often prohibitively expensive because they require complex formulations and large numbers of simulations. ML has been employed as a robust and powerful surrogate in the required forward simulations and thus become a popular toolbox in inverse problems~\cite{xu2020inverse,yang2021b,meng2020composite,zhou2021markov,mcbane2021component,choi2020gradient,choi2019accelerating,amsallem2015design}. Recent advances are also made towards forward problems from the rising research interests on physics-informed neural networks (PINNs)~\cite{raissi2019physics,lu2020extraction}, which complement NN training with physics knowledge.

\subsection*{Knowledge gaps and challenges}
\addcontentsline{toc}{subsection}{Knowledge gaps and challenges}
\paragraph{\underline{Data vs. Physics}} The first challenge is about data. Classical ML arises from computer science, in applications, for example, image classification, face recognition, and motion detection, where ``big" data are available and free of charge to the analyst. Classical ML ``learns" information directly from data without relying on physical laws or predetermined equations because it is mostly driven by problems for which no physical laws or predetermined equations are available. The concept of prediction is more qualitative than quantitative. As far as classical ML is concerned, there is no inherent difference between images of cats and temporal snapshots of, say, a contaminant plume migrating through a subsurface environment.

Subsurface data are usually “small” data, meaning that there are scarce data to capture all phenomena, especially for real large-scale problems. Physical systems usually consist of a small amount of available data, which are either difficult or expensive to obtain. Conventional physics-based modeling provides computationally expensive simulation data, but the concepts of error and accuracy are universal, and the concept of prediction is mostly quantitative. More recently, physics-informed data-driven modeling has been studied to tackle this challenge, as discussed later in the subsection ``How to incorporate physics in ML".

Nowadays, with the development of data science, modeling for the most challenging multi-scale, multi-physics, complex systems has been driven to an intermediate area, where the models are constructed by some data and some physics. Even if we have access to more data with advanced computing capacity, the success of ML relies heavily on our ability to collect and interpret data. Due to the multi-scale and multi-physics nature of the problem, synthesizing and creating data via numerical tools can be computationally expensive. Numerical algorithms are still under development for efficient preservation of relevant physics, for example, shock-capturing schemes in a sharp interface, asymptotic-preserving schemes in different scales, mass/momentum/energy-preserving schemes for conservation laws, symplectic structure-preserving for Hamiltonian systems. The so-called ``curse of dimensionality" addresses the challenge of data in high dimensional systems and stochastic systems. There are more difficulties in dealing with experimental data, including cleaning the data from noisy measurements, cooperating the data in the right scales, addressing observation bias, and interpreting modeling errors.

The extensive use of data-driven modeling is fundamentally transforming science and engineering, especially in the areas where accurate or computable models remain elusive. Although there were examples of constitutive laws ``re-discovered" by ML (e.g., Darcy’s law), how to cooperate multiscale modeling with ML learning of constitutive laws remains unclear. ML can not do better than conventional homogenization tools in multiscale systems. Given a dataset of all scales, ML, without a good understanding of the physics,  still can not figure out the dominant scales and simplify the problem for us in a systematic way. We should keep the famous saying ``all models are wrong, but some are useful" in mind. Data-driven modeling and physics-based modeling should complement each other depending on the data availability and application scenarios. The task to bridge the gap between data-driven modeling and physics-based modeling remains a major challenge to this very day.

\paragraph{\underline{Forward Problem vs. Inverse Problem}}
Solving inverse problems, e.g., inferring material properties or identifying transport sources, is often prohibitively expensive using traditional numerical solvers because it requires complex formulations and large numbers of simulations. Machine learning has been employed as a robust and powerful surrogate in the required forward simulations and thus become a popular toolbox in inverse problems~\cite{xu2020inverse,yang2021b,meng2020composite,zhou2021markov,mcbane2021component,choi2020gradient,choi2019accelerating,amsallem2015design}.  Within the prior distribution, machine learning is extremely good at interpolation. Many generative approaches, including variational autoencoder (VAE) and generative adversarial networks (GANs), have been widely used in generating data within predetermined distributions.

However, machine learning in solving forward problems is rather challenging~\cite{fuks2020limitations,wang2021and,fraces2021physics}. This is partially due to the lack of fundamental mathematical theories for neural network (NN) approximations. Our understanding of NN approximations stays at the universal approximation theorem~\cite{hornik1989multilayer}. However, the studies on convergence and error bound remain unclear and thus fall far behind the rapid developments of variant NN structures. Recent work uses the underlying discretized equations to derive an error bound in the context of ROMs~\cite{kim2021fast}. Studies on meta-learning~\cite{psaros2021meta,chen2021learning} and  automated machine learning (autoML)~\cite{he2018amc,he2021automl,gijsbers2019open} are under development. Without a principled way of selecting structures and training NNs, it is hard to validate NN-based surrogate models and trust the predictions for a long time as we do in numerical simulators that are equipped with well-studied stability and accuracy. It is also common that NN-based surrogates are very problem-specific -- they are not well adapted to different boundary/initial conditions and can not be generalized to out-of-distribution inputs. There are very few works (e.g., \cite{ruthotto2020deep}) on the relationship between NN structure and specific problems. Moreover, the outputs of the NN-based surrogates are not physically explainable in many multi-scale problems. As an analogy to conservative schemes in numerical simulators~\cite{leveque1992numerical}, a carefully designed NN structure with physics ingredients in conservation laws is needed to fully capture and interpret the physics in the systems. In particular for multiscale problems, it has been realized that NN has difficulty learning high-frequency functions (known as ``F-principle"~\cite{xu2019frequency} or ``spectral bias"~\cite{rahaman2019spectral}). Until recently, advances in learning high-frequency functions have been achieved via a phase shift~\cite{cai2020phase}. Although machine learning has been successfully used in surrogate modeling, many unanswered questions remain about accuracy, generalizability, and interpretability, even for the most simple problems.

\subsubsection*{\underline{Example: Riemann problem using Physics Informed Neural Networks (PINN)}}
\begin{figure}[htp!]
    \centering
    \includegraphics[width=0.8\linewidth]{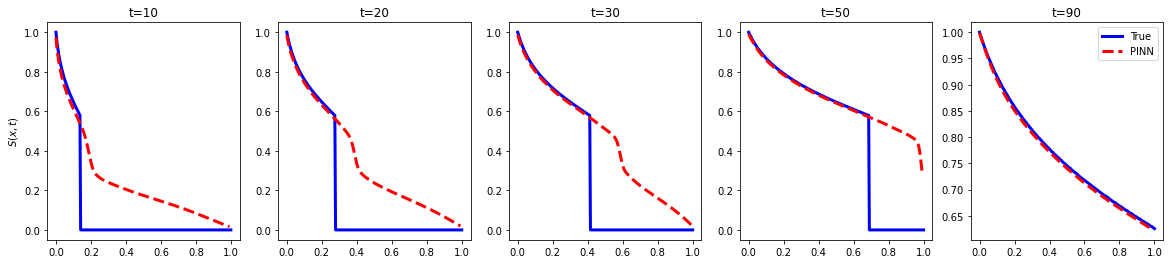}
    \caption{Results of saturation inference computed using the PINN approach (dashed red) conditioned on the weak form of the hyperbolic problem. The reference solution using the method of characteristics (MOC) is plotted in blue.}
    \label{fig:BL_saturation_fail}
\end{figure}

\addcontentsline{toc}{subsubsection}{Example: Riemann problem using Physics Informed Neural Networks (PINN)}
It is possible to solve challenging forward problems using deep neural networks. A popular approach is Physics Informed Neural Networks (PINN) --  neural networks trained to solve supervised learning tasks while respecting any given laws of physics described by general nonlinear partial differential equations. We present an example where an immiscible two-phase flow displacement (Buckley-Leverett problem) in a heterogeneous porous medium is modeled using such a model. Various attempts had been made at solving this hyperbolic problem with non-linear, non-convex flux function with multiple inflections (Eq.~\ref{eq:residual_buckley} and \ref{eq:frac_flow}). The direct and simple application of PINNs presented in ~\cite{raissi2017physicsI} does not allow to produce satisfactory solutions (Figure~\ref{fig:BL_saturation_fail}). For reference, this method uses a neural network to solve partial differential equations (PDE) of the form:
\begin{equation}
    \mathcal{R}(\mathbf{X}, t, \mathbf{\nu}, \mathbf{\nu}_t, \nabla\mathbf{\nu}, \nabla^2\mathbf{\nu},\dots) = 0
\end{equation}
In transport problems, $\mathbf{\nu}$ typically refers to state variables such as the fluid saturation or concentration $\mathbf{\nu}=S_{o,g,w}$.
We assume that the unknown solution $\nu$ is represented by function compositions with series of linear and non linear transformations. This can conveniently be represented by a feed forward multi-layer perceptron.
\begin{equation}
    \mathbf{\nu} \approx \mathbf{\nu}_{\theta} = \sigma\left[\mathbf{W}^{[n]}\times \sigma(\mathbf{W}^{[n-1]}(\dots\sigma(\mathbf{W}^{[0]}[\mathbf{X},t]^T + \mathbf{b}^{[0]}))\dots + \mathbf{b}^{[n-1]}) + \mathbf{b}^{[n]}\right]
    \label{eq:ffwd_form_base}
\end{equation}
Where $\sigma$ is a nonlinear activation function (sigmoid, tanh, ReLu,...), $\mathbf{W}_i$ are weight matrices and $\mathbf{b_i}$ are bias vectors for layer $i$. 

$S_{w,\theta}(x,t)$ is a continuous representation of the state variables. For any $x,t$, it will produce an output. It is also a differentiable representation. This means that we can compute any derivatives of it with respect to space and time. We can therefore construct the new quantity of interest $\mathcal{R}$ (for residual). We leverage the capabilities of modern software libraries like \textit{Tensorflow} to compute the gradients in the PDE via automatic differentiation.

A recent analysis~\cite{fuks2020limitations} demonstrates the need for further theoretical and algorithmic developments of PINN and other machine learning techniques for solving real-world problems. The study focused on nonlinear one-dimensional hyperbolic problems with various flux functions. It found severe limitations of the off-the-shelf implementation of PINN in solving the Riemann problem with shocks and discontinuous (mixed wave) solutions. An example of saturation front resulting from such resolution is presented in Figure~\ref{fig:BL_saturation_fail}.

This result conforms with the one presented in~\cite{fuks2020limitations} and challenges the idea that NN-based strategies can be used to solve \emph{any} PDE. It highlights the need for both theoretical and algorithmic developments aimed at addressing the challenges associated with scalability and the presence of computational bottlenecks; possible venues to explore are information-theoretic approaches, domain decompositions, and probabilistic graphical models~\cite{taverniers-2021-mutual,hall-2021-ginns}. Crucially, such results demonstrate the need to combine the domain-specific knowledge with NN-based strategies rather than replacing the former with the latter.

For example, in~\cite{fuks2020limitations}, the limitations of the off-the-shelf version of PINN were overcome by bringing additional physical constraints to the formulation of the problem. We remind that the Buckley-Leverett problem characterizing the transport of a two-phase system in porous media can be solved in one dimension using various methods. The original work of Buckley and Leverett~\cite{buckley1942mechanism} dealt with an equation
\begin{equation}
\label{eq:residual_buckley}
    \frac{\partial S}{\partial t} + \frac{\partial f(S)}{\partial x} =0
\end{equation}
where the fractional flow $f$ is a nonlinear equation defined as:
\begin{equation}
    \label{eq:frac_flow}
    f(S) = \frac{(S - S_{wc})^2}{(S - S_{wc})^2 + (1 - S - S_{gr})^2/M}
\end{equation}
subject to constant boundary and initial conditions:
\begin{eqnarray}
    S(x=0,t) &= S_{inj}\\
    S(x,t=0) &= S_{wc}
    \label{eq:BL_uniform_bc}
\end{eqnarray}
where $S_{wc}$ represents the residual (connate) saturation of the wetting phase (typical water), $S_{gr}$ the residual saturation of the non-wetting phase, and $M$ the endpoint mobility ratio between the two phases defined as the ratio of endpoint relative permeability and viscosity of both phases.
The resolution of the partial differential equation \eqref{eq:residual_buckley} in its weak form, along with classical Dirichlet initial and boundary conditions, leads to a non-physical solution. Indeed, the saturation obtained can be doubled (or tripled) valued at a given $x$ and $t$.
Various methods have been proposed (\cite{welge1952}, \cite{buckley1942mechanism}) to resolve the apparent inconsistency between the weak form of the PDE described in Eq.~\eqref{eq:residual_buckley} and conservation law that as an integral law may be satisfied by functions that are non-differentiable. Lax (\cite{Lax_Hyperbolic}) provides a detailed theory of conservation law featuring shock waves. This framework gives rise to the \emph{principle of entropy} that stipulates the entropy of particles crossing a shock front must increase. 
Orr~\cite{Orr_gas_injection} develops a general theory of multi-component, multi-phase displacement based on the MOC. The Buckley-Leverett solution can be seen as a special case with two phases and one component per phase.
The theory shows that the differential form of the PDE along with necessary boundary and initial conditions are not sufficient to get a physically consistent solution to the problem. Two physical principles constraining the wave velocities are introduced to address the problem:
\begin{enumerate}
    \item The \emph{Velocity constraint}: Wave velocities in a two-phase displacement must decrease monotonically as the upstream saturation continuously increase.
    \item The \emph{Entropy condition}: Wave velocities on the downstream side of the shock must be less or equal to the shock velocities and the velocities on the upstream of the shock.
\end{enumerate}
These two physical constraints are resolved by forcing the displacement to occur on the convex hull of the fractional flow.
We demonstrate that the PINN approach must be bound by the same set of physical constraints in order to produce a physically consistent solution. Therefore we must replace the original fractional flow equation with one describing the convex hull. The parameterization is defined as:
\[
  \tilde{f}(S) =
  \begin{cases}
                                   0 & \text{if $S\leq S_{wc}$} \\
                                   \frac{S - S_{wc}}{f(S_f)} & \text{if $S_{wc}\leq S\leq S_f$} \\
  \frac{(S - S_{wc})^2}{(S - S_{wc})^2 + (1 - S - S_{or})^2/\tilde{M}} & \text{if $S>S_f$}
  \label{eq:frac_flow_welge}
  \end{cases}
\]

\begin{figure}[htp!]%
    \centering
    \includegraphics[width=1\linewidth]{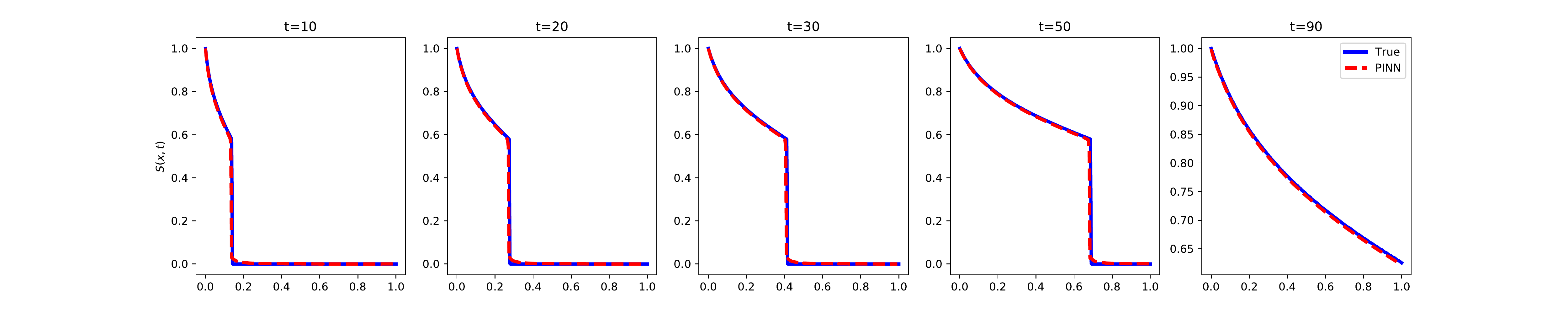}%
    \caption{Results of saturation inference using PINN (dashed red) vs. MOC (blue) with velocity constraint and entropy condition. The convex hull of the fractional flow curve is used to model the displacement.}%
    \label{fig:BL_uniform_fit}%
\end{figure}

We differentiate the piecewise function $\tilde{f}(S)$ with respect to $x$ in order to compute the residual (Eq.~\eqref{eq:residual_buckley}). The results of the simulation are presented in Figure~\ref{fig:BL_uniform_fit}.

We observe a close match with the reference solution, including around the shock. The $L_2$ error after $30,000$ iterations is approximately $2\times 10^{-3}$.

\begin{figure}[htp!]%
    \centering
    \includegraphics[width=1\linewidth]{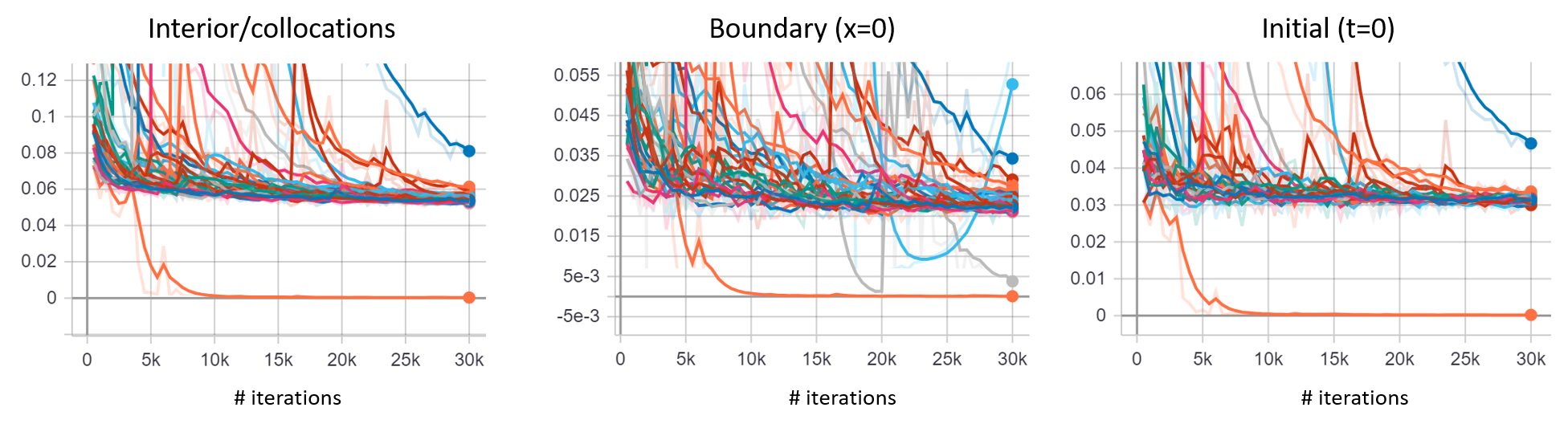}%
    \caption{Evolution of $L2$-loss with number of iteration for: collocation points (left), boundary condition ($x=0$) and initial condition ($t=0$). All curves except for the bottom (orange) one correspond to experiments with the network architecture. The orange one corresponds to the addition of entropy condition to the training.}%
    \label{fig:BL_loss_comparison}%
\end{figure}

The result of this study shows that the addition of a physical constraint in line with computational flow dynamics theory is necessary to solve the conservative transport problems using PINNs. On the other hand, variations of the hyperparameters and model architecture do not have a conclusive impact on the accuracy of the solution. We substantiate this by plotting the evolution of the loss functions for 164 experiments run on a variety of network architectures (Deep Galerkin, Fourier, Residual network with various layer sizes and activation functions) and overlay the loss corresponding to Figure~\ref{fig:BL_loss_comparison} for reference.

The loss function comparison shows that all the deep learning architectures tested did meet some limitations in reducing training loss. We can see the formation of a lower ``plateau" for all the loss functions but one.
On the other hand, we observe a clear improvement when the entropy condition is added to the problem. This result indicates that substantial improvements in the applicability of PINNs can be achieved through physical reasoning rather than through pure network parameters tuning.

\paragraph{\underline{Nonlinear approaches vs. Linear approaches}}
Nowadays, model reduction techniques play a more and more important role in compute-intensive science and time-critical applications. For example, digital twins, a virtual representation that serves as a
real-time reduced-order model of a physical process. It aims to calibrate the computational models with measurement
data and enable real-time decision-making and control. Applications of digital twins include energy storage system design, virtual reality games, robotic surgery, and many more. It is an important part of data-driven modeling because efficiency is why we shift from expensive physics-based modeling to ``cheap" data-driven modeling. Its importance has been recognized and advocated by governments and industries as well.

The model reduction field was dominated by linear approaches and projection-based method~\cite{benner2015survey}. Linear methods are simple and fast. Also, it is easy to construct extrapolations from linear models as well as make analyses from eigenvalues and dominant modes. The counterpart of the physical quantities can be found in the reduced dimension, and structure preservation can be carefully designed. Error analysis and model interpretability are both well-established for linear approaches. Unfortunately, most physical phenomena are nonlinear systems or with Kolmogorov width decaying slowly(e.g., advection-dominated problems) and thus can not accurately be represented by a low dimensional linear model. There are several linear subspace ROMs which have been successfully applied to nonlinear systems~\cite{carlberg2018conservative,choi2019space,kim2020efficient,choi2020sns,hoang2021domain,copeland2022reduced}.

On the other hand, modern nonlinear methods like NN, with various architectures and flexible parameters, have become extremely powerful approximators in representing data and interpolating between data. As mentioned in the previous subsection, generalizability in the extrapolation regime is a challenge for NN due to our limited knowledge about NN approximation theory. The huge number of NN parameters also leads to the so-called ``overfit" issues. As always, we should be aware of the trade-off between accuracy and computational costs (both online and offline included). The choice of traditional or modern methods should depend on the problem and application.

\subsection*{Recommendations to advance the field in ten years}
\addcontentsline{toc}{subsection}{Recommendations to advance the field in ten years}
Faced with the challenges mentioned above, we are also presented with tremendous opportunities. We would like to advance the field in ten years with the following questions in mind and hopefully answer them with new understandings.

\paragraph{\underline{How to generate and use data?}}
Improving numerical simulators is still a significant research topic. Users, funding agencies, and journal reviewers should not underestimate the value of efficient numerical simulators even if the speedup is only a factor two. Industrial calibration should also be included in the evaluation as a consistent 10\% improvement in efficiency is considered important in practice. An efficient numerical simulator also plays an important role in reducing the computational costs in ML training, which should not be underestimated.  
 
\begin{figure}[h!]
     \centering
     \includegraphics[width = \textwidth]{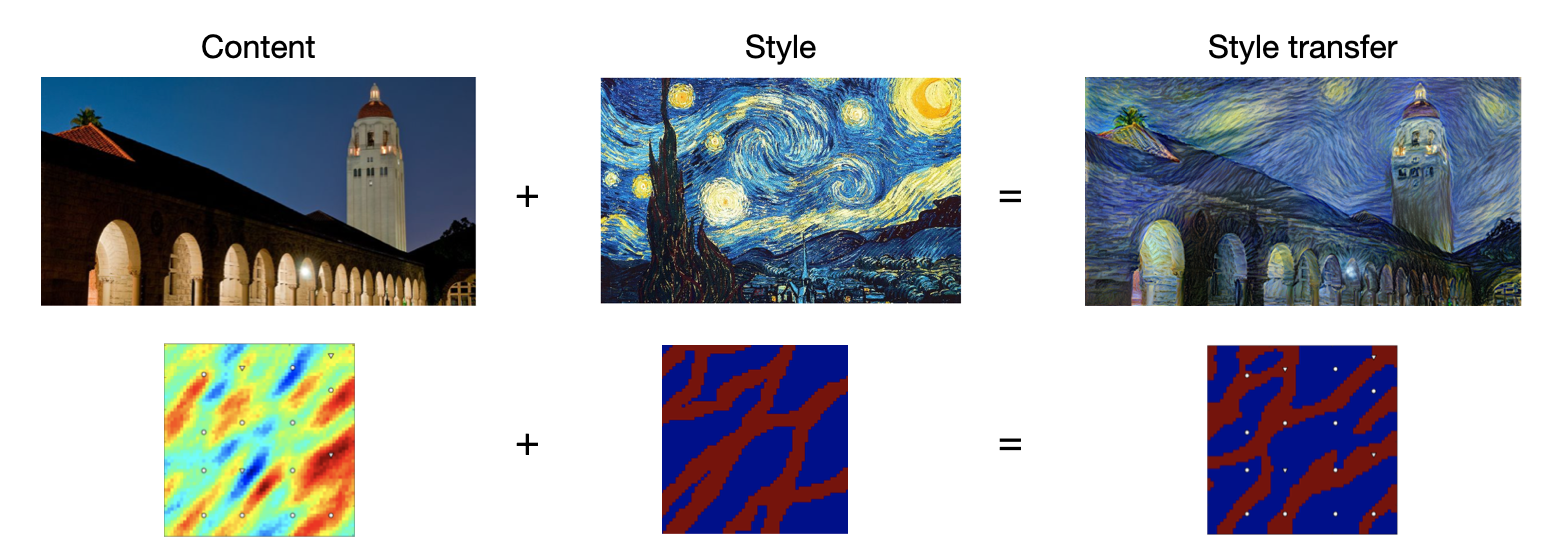}
     \caption{Top: image generation via transfer learning; Bottom: channelized permeability map generation via transfer learning~\cite{cs231n}.}
     \label{fig:TransferLearning}
 \end{figure}

At the current stage (and possibly in the following ten years), it is not possible to completely replace conventional numerical simulators with ML-based surrogates. However, ML can be used to facilitate the generation of data, which could enable learning from disparate data (a hierarchy of data sources instead of a single data source). For example, a popular way of generating more data in computer science is via transfer learning, where the input content and desired style gives a Van Gogh style Hoover Tower in Figure~\ref{fig:TransferLearning}, which is drawn efficiently by ML instead of Van Gogh. Applications of transfer learning also have been used in scientific computation as well~\cite{tang2020deep}. Given a connected, channelized permeability map (which is usually very expensive to generate) as the desired style and a low dimensional Principal Component Analysis (PCA) content, transfer learning can help to enforce the channelized, connected features on the low dimensional PCA representative and provide an efficient generation of data with desired features.
 
Another perspective is how to sample data. Adaptive importance sampling~\cite{rotskoff2020active} is needed to sample rare events that dominate the loss function, which could require a certain kind of non-local sampling based on flow map. In rare-event regimes where very few actual data samples are available, physics-informed ML could be used to complement the shortage of data. This combination can reduce the asymptotic variance of the solution and improve generalization. There are many more open questions in this area, including selecting ML architectures in adapting to a given system.

\paragraph{\underline{How to incorporate physics in ML?}}
From our examples and discussion, we conclude that the future directions should focus on how to make physics-based models and ML-based surrogates complement each other instead of one replacing the other.

There are rising research interests on physics-informed neural networks (PINNs)~\cite{raissi2019physics,lu2020extraction}, which complement NN training with physics knowledge. In terms of homogenization, there are also recent works on using physics knowledge and ML to couple different scales more efficiently. Using ML to improve existing physics laws is more promising than discovering new laws at the current stage. PDEs, constitutive relations, experimental data, and ML should be integrated to give us the governing laws. Computational scientists and experimentalists should join forces to achieve advances in mechanical design and control.

Many open questions remain, especially for high-dimensional, nonlinear dynamical systems that exhibit rich multiscale phenomena in both space and time. However complex, many of these systems have well-studied properties in existing theoretical and experimental research, which may serve as a promising tool
in efficiently identifying the low-rank structure of the underlying dynamics. Recent work of Koopman theory-based methods~\cite{kutz2016dynamic,lu2020lagrangian,lu2021dynamic,lu2020prediction} serves as a bridge between nonlinearity and linearity, bringing in new perspectives in model reduction and ML interpretability. Similar advances are needed to address the challenges of machine learning interpretability and incorporation of domain knowledge. 

There are tremendous opportunities and challenges in this emerging research area of high-dimensional input data with noise, the convergence of the algorithms, selection of neural network architectures, the generalizability of the model, to name a few. Statistical tools and numerical analysis for uncertainty quantification and probabilistic modeling also need to be incorporated in the learning process to handle some of these challenges.

\paragraph{\underline{Deterministic approaches vs. Probabilistic approaches}}
Uncertainty quantification (UQ) plays a vital role in reservoir engineering problems. Traditional Monte Carlo based methods can be prohibitively expensive due to a large number of simulation runs required. In the following context, we present an example of using the advantages of neural networks to interpolate in a high dimensional space in UQ tasks.

We present a parameterization of the Physics Informed Neural Network (P-PINN) approach to tackle the problem of uncertainty quantification in reservoir engineering problems. We demonstrate the approach with the immiscible two-phase flow displacement (Buckley-Leverett problem) in a heterogeneous porous medium. The reservoir properties (porosity, permeability) are treated as random variables. The distribution of these properties can affect dynamic properties such as the fluids' saturation, front propagation speed, or breakthrough time. We explore and use to our advantage the ability of networks to interpolate in high dimensional space. We observe additional dimensions resulting from a stochastic treatment of the partial differential equations tends to produce smoother solutions on quantities of interest (distributions parameters) which is shown to improve the performance of PINNs. We are able to solve problems that can be challenging for classical methods. This approach gives rise to trained models that are both more robust to variations in the input space and can compete in performance with traditional stochastic sampling methods.  

This example is based on stochastic analysis of the Buckley-Leverett problem~\cite{Zhang_tchelepi1999} in which the rock properties (porosity, permeability) are non-uniform and distributed randomly across space. This leads to the conservation law derived in Eq.~\ref{eq:conservation_vd}:

\begin{equation}
    \label{eq:conservation_vd}
    \frac{\partial S_w}{\partial t} + v_d f_w^{\prime}(S) \frac{\partial S_w}{\partial x} = 0
\end{equation}

where $v_d=q(x)/\phi(x)$ is the total (Darcy) velocity vector. Since $\phi$ is a random function, this has an effect on $v_d$ and $S_w$ which are treated as a random variables. We will explore the effect of varying the distribution parameters for $v_d$ on $S_w$.

We describe the formalism of the physics informed approach applied to the stochastic problem. We remind that for the deterministic problem, the input of the model is the time and space dimension, and the output is the saturation. In the stochastic version, additional dimensions are added to the input sampling space and enable to produce a single neural network model over the entire uncertainty range. We conveniently exploit the ability of the neural network to interpolate in high dimensions. 
Just like in Eq.~\eqref{eq:ffwd_form_base}, the model is a dense neural network of the form:
\begin{equation}
    \mathbf{\nu} \approx \mathbf{\nu}_{\theta} = \sigma\left[\mathbf{W}^{[n]}\times \sigma(\mathbf{W}^{[n-1]}(\dots\sigma(\mathbf{W}^{[0]}[\mathbf{X},t, v_d]^T + \mathbf{b}^{[0]}))\dots + \mathbf{b}^{[n-1]}) + \mathbf{b}^{[n]}\right].
\end{equation}

\begin{figure}[h!]
    \centering
    \includegraphics[width=0.8\linewidth]{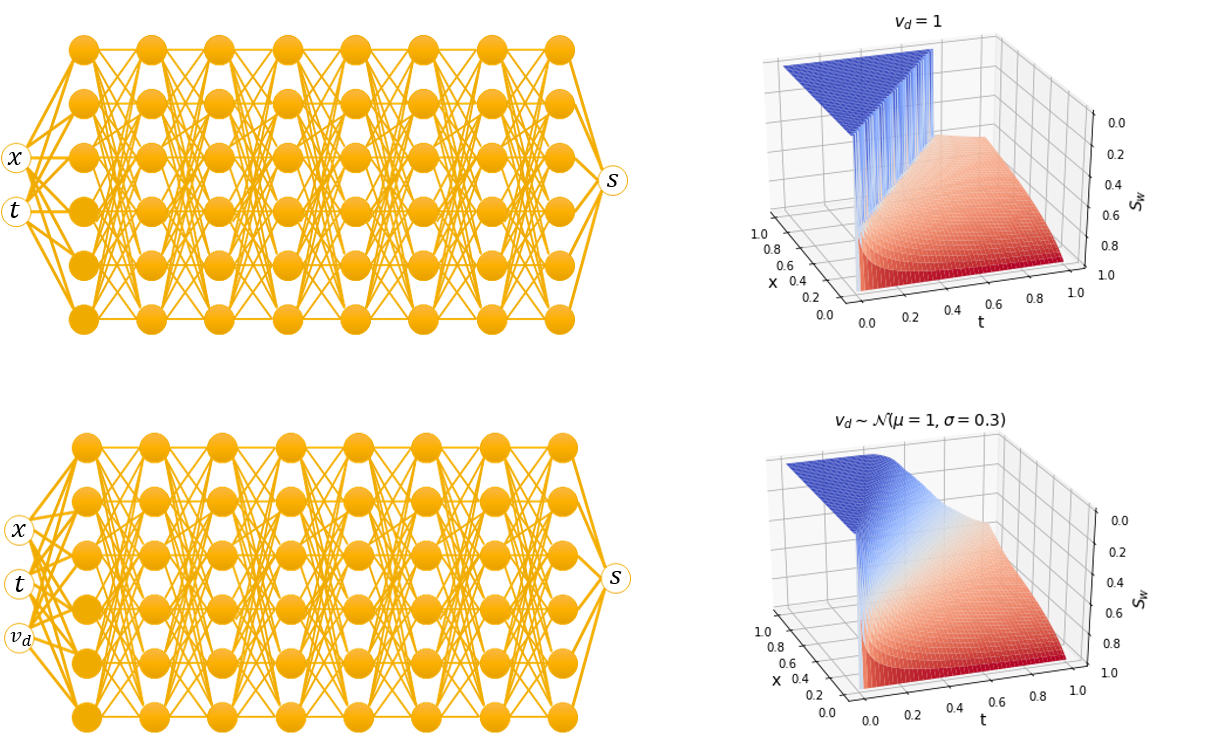}
    \caption{Presentation of the model (input/output) and solution for the deterministic (top) and probabilistic (bottom) Buckley-Leverett problem.}
    \label{fig:mlp_probabilistic}
\end{figure}

Figure~\ref{fig:mlp_probabilistic} shows the difference between the deterministic and stochastic models and solutions.

We see that while the deterministic solution requires a lower-dimensional input space, it features sharp gradients (discontinuity, shocks). The stochastic solution, on the other hand, is smoother. We present the one-dimensional Buckley Leverett solution with total velocity $v_d$ constraint where the velocity is constant for a given realization but varies from one realization to the next. With a single training using the parameterized PINN (P-PINN) approach, we intend to produce a solution that will honor solutions for all realizations within the bounds of the distribution initially used. This formulation allows exploring the stability of PINNs to variations in the input space. This characteristic of stochastic solutions is interesting for training as we know that solutions with steep gradients can be challenging when solving PDEs using PINNs.

We present results where the Darcy velocity $v_d$ follows a one-dimensional-Gaussian distribution:

\begin{equation}
    \label{eq:distrib_v_d_narrow}
    v_d\sim \mathcal{N}(\mu=1,\,\sigma=0.3, \text{low}=0.5, \text{up}=2).
\end{equation}

We truncate the distribution between $0.5$ and $2$ for physical consistency (positive velocity) and to ensure that the shock front remains within spatial bounds for most of the simulation duration.

\begin{figure}[htp!]%
    \centering
    \includegraphics[width=1\linewidth]{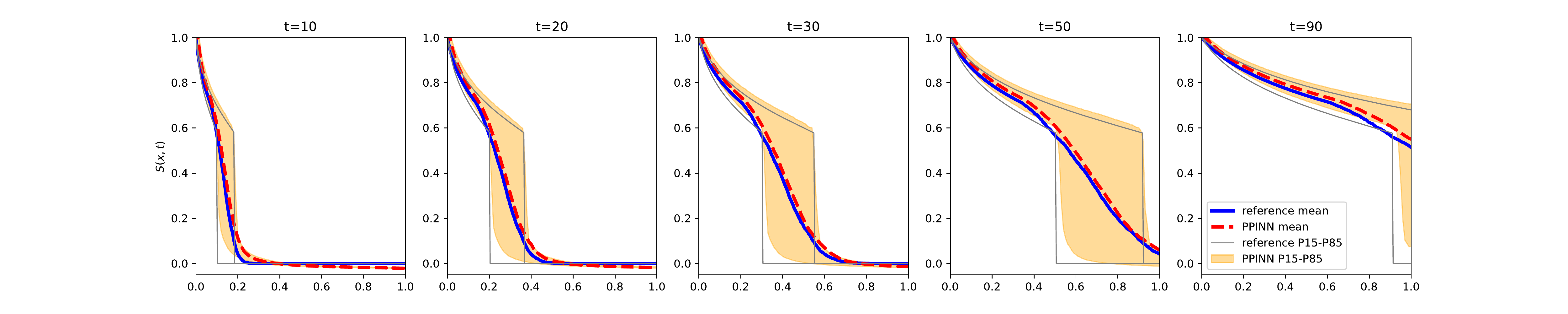}%
    \caption{Comparison of saturation distributions profiles at five different time steps for a Total velocity $v_d\sim \mathcal{N}(\mu=1,\,\sigma=0.3, \text{low}=0.5, \text{up}=2)$. The reference mean saturation (computed through Monte Carlo simulation of MOC) is solid blue, while the one computed with P-PINNs is dashed red. The P15-P85 envelopes are represented for both reference and P-PINN.}%
    \label{fig:sat_dist_vd_narrow}%
\end{figure}

Figure~\ref{fig:sat_dist_vd_narrow} shows a comparison of the probabilistic solution obtained with the P-PINN. We represent the mean of all the realizations for the P-PINN approach. The envelope around the profile represents the $15$th to $85$th percentile. It is compared with a Monte Carlo simulation on the MOC with $1000$ samples.

\begin{figure}[htp!]%
    \centering
    \includegraphics[width=1\linewidth]{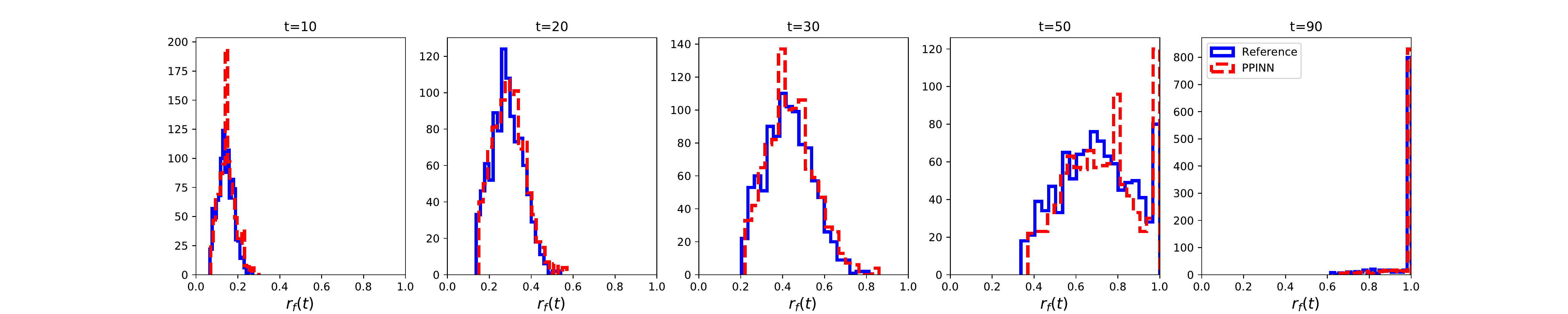}%
    \caption{Comparison of front radius distributions profiles at five different time steps for a total velocity $v_d\sim \mathcal{N}(\mu=1,\,\sigma=0.3, \text{low}=0.5, \text{up}=2)$. The reference front radius (computed through Monte Carlo simulation with MOC) is solid blue while the one computed with P-PINNs is dashed red.}%
    \label{fig:front_radius_vd_narrow}%
\end{figure}

The Buckley-Leverett model is typically used to predict the progression of the front for the injected fluid along with its breakthrough time at given locations. The distributions of shock radii are shown in Figure~\ref{fig:front_radius_vd_narrow}. They present close matches with each other.

This example illustrates that quantities of interest (QOI's), such as front radius, can be well captured provided that we establish a proper parameterization of the input uncertainty space. We observe that the saturation distributions at specific space and time coordinates are well matched for a homogeneous case. A possible extension of this method could be to approximate the output probability function directly by solving differential equations for the moments of the distribution \cite{zhang2001stochastic}.

\paragraph{\underline{How to balance accuracy and expressiveness?}}
New challenges arise from the need for robust and reliable predictive ML tools for physical systems. Despite many empirical successes of ML, little is known to us about the theoretic foundations of NN approximation. How to validate the ML-based models and how to build trust in the real-time decisions made by ML will determine the future of artificial intelligence. We are looking for mathematical and statistical developments in approximation theory and optimization. Fortunately, we can learn from and build on our experience of finite elements. The concepts of stability and convergence in traditional numerical analysis are expected to be further explored in the context of ML in the next ten years. 

Meanwhile, with the rapid development of GPU capacities (almost double every year), the speedup by ML surrogates is not the most important feature in the following years. Understanding the physics and mechanics from ML surrogates is the new challenge. Discovering physical laws is often achieved by dictionary learning~\cite{bakarji2021data}, where dominant operators are determined from data. How to propose the dictionary requires better knowledge about how to define the problem in the right scale. Overloading the dictionary with unnecessary terms not only slows down the learning process but also produces to misleading identifications. Therefore, model accuracy is not well defined if the scale of interest is not well defined. We hope that researchers from domain knowledge background can take the lead in defining the right problems for ML algorithms so that the error can be rigorously defined and the model can be properly interpreted.

In summary, our state-of-the-art review and discussion support the general trend that machine learning can play a critical role in understanding multi-scale phenomena in heterogeneous porous materials. Our examples suggest that it will be critical to integrate data, physics, and machine learning tightly. A multidisciplinary approach will be crucial to develop, test, and validate reliable computational tools to achieve this goal. Physics-informed neural networks are a powerful and promising first step in this direction.

\addcontentsline{toc}{subsection}{References}
{\renewcommand{\markboth}[2]{}% Remove header adjustment
\printbibliography}
\newpage
\end{refsection}
%TA2 - ML in porous and fractured media
\begin{refsection}
\begin{center}
\section*{ML in porous and fractured media}
\addcontentsline{toc}{section}{ML in porous and fractured media}
\author*{Ma\v{sa} Prodanovi\'{c}$^1$, Bo\v{z}o Va\v{z}i\'{c}$^2$,  Daniel O'Malley$^3$, Hang Deng$^4$, Hongkyu Yoon$^5$ and Gowri Srinivasan$^3$}
\end{center}
%Overall the entire section should be 5000 words with 2 figures

\noindent
$^1$ The University of Texas at Austin, TX, USA.\\
$^2$ The University of Utah, UT, USA.\\
$^3$ Los Alamos National Laboratory, NM, USA.\\
$^4$ Lawrence Berkeley National Laboratory, CA, USA.\\
$^5$ Sandia National Laboratories, NM, USA.

\subsection*{Abstract}
\addcontentsline{toc}{subsection}{Abstract}
Scalable and predictive modeling for flow and transport through natural porous and fractured media, as well as their mechanical behaviors, is essential for a variety of engineering and science applications, including, but not limited to, groundwater management, energy resources recovery, $\text{CO}_2$ sequestration, and contaminant remediation. Predictive modeling in geological systems is challenging, partly, due to the varying degree of structural heterogeneity and compositional complexity. State of the art in multi-scale modeling was covered in the previous chapter. In this chapter we will focus on how machine learning (ML) can help characterize fractured and porous media and examples of modeling coupled processes. For example, the length scales of individual pores and grains (nm-mm length scale) often control the transport (e.g., trapping of $\text{CO}_2$) or mechanical behavior (e.g., crack initiation at weak grain contacts), resulting in spatio-temporal multiscale phenomena which can only be accurately modeled at a high computational cost. Multiscale aspects in geological systems are also ubiquitously common in many other manufactured materials such as membranes, fibrous materials, solid foams, and fuel cells. Machine learning (ML) has recently advanced as a powerful tool for accelerating these computationally expensive simulations and for interpreting the essential features that control model behavior. 

Here, we review the state-of-the-art in machine learning applied to physical modeling in porous and fractured media, existing data, and benchmarks that can serve for ML training, and the challenges in broadening participation in this research community. We conclude that the existing and quite advanced modeling tools need to be complemented by easier methods for sharing, as well as easier access to curated data and high-performance computing is required to perform simulations and ML applications efficiently.   

\subsection*{Introduction}
\addcontentsline{toc}{subsection}{Introduction}
In porous and fractured media, flow and transport are critical for multiple applications, including energy security and storage material design. Predictive models are valuable tools for uncertainty assessments and risk-informed decision-making. Topics of interest in natural systems include modeling of the geological process of rock formation and subsurface engineering applications such as management of groundwater resources, carbon sequestration, hydrocarbon extraction, and contaminant transport. In manufactured porous media and relevant materials science, applications include, e.g., improving the absorption capacity of diapers, the energy capacity of storage materials, or shock-absorbing properties of solid foams. We often have to characterize the media and model complex physics of interest at multiple length scales in these applications.

To properly integrate pore-scale physics into field-scale predictions, it is critical to collect pore-scale structural information and quantify local properties and physical processes, typically via high-resolution experimental measurements and/or high fidelity modeling. Both tasks are complex and time-consuming and often have to account for spatio-temporal dynamic behavior and work with either too large or too sparse datasets. 

\begin{figure}[!h]
    \centering
    \includegraphics[scale = 0.425]{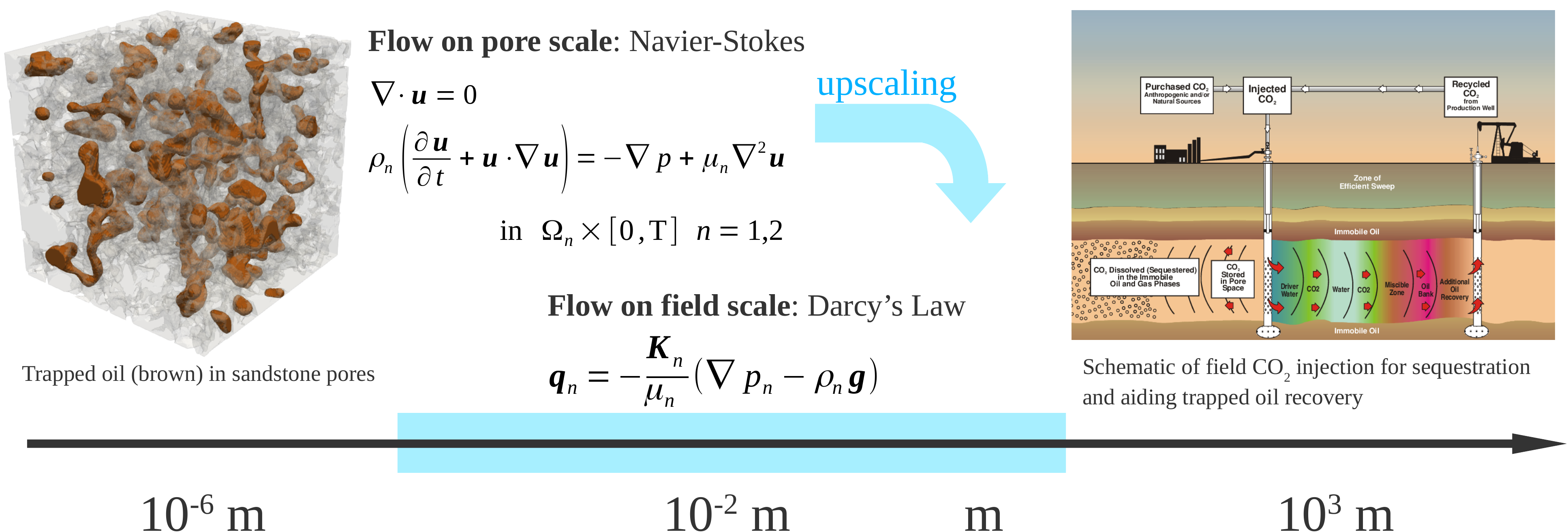}
    \caption{An example range of length scales in a carbon storage/utilization application. The pore-scale capillary forces dictate trapping as shown in the image on the left (visualized based on \cite{victor2016minimum}) and affect the properties, such as permeability to each phase $n$, labeled $K_n$ that are required as input in field-scale simulations. The schematic of field scale operations is reprinted from \cite{lake2019co2}.}
    \label{fig: intro}
\end{figure}

Figure \ref{fig: intro} shows an example of two-phase flow in porous media where at the pore-scale, each phase and boundaries are distinctively identified, and blobs of the non-wetting fluid (e.g., oil in the reservoir or supercritical carbon dioxide at the post-injection) are trapped. This pore-scale process is predominantly controlled by local capillary forces that largely depend on pore sizes and mineral properties (i.e., wettability). In modeling reservoir- and aquifer-scale behavior of systems such as geologic carbon storage reservoirs, trapping is typically accounted for through constitutive relations between saturation and relative permeability or capillary pressure, in which important pore-scale heterogeneity and processes are simplified or completely lost. ML methods have been particularly versatile in complex pattern recognition, discovering new equations (Chapter 1), and providing surrogate models for complex systems. Therefore, they can be extremely valuable to advancing our ability to extract key features from complex, sparse, or big data, to identify primary mechanisms of coupled processes for data analytics, to develop physics-constrained reduced-order modeling for real-time parameter estimation and uncertainty quantification, and to predict emergent phenomena that can be explored via integrated data analytics and physics-based learning for new scientific discovery.

Flow and transport in fractured media suffer from much of the issues mentioned for porous media, with the additional complexity of striking a balance between simplifying and homogenizing models for computational tractability while capturing the underlying structure of the fracture network. Additionally, flow and transport in fractured media occur only on a small subset of the entire domain, which introduces large errors when applying modeling approaches that resort to homogenizing or averaging properties. In fact, the topology of the fracture network plays a key role in determining transport characteristics and cannot be ignored relative to the geometric characteristics or transport/hydrologic properties. For example, system behavior is very much dependent on whether large fractures intersect or are connected by smaller fractures. However, such structural information cannot be fully characterized at the macroscale due to the high computational cost incurred in representing the discontinuities formed by cracks that demand highly resolved meshes. While there are high fidelity mesh-based fracture models capable of representing millions of micro-fractures, such as dfnWorks \cite{noauthor_dfnworks_nodate} and HOSS \cite{noauthor_HOSS}, the computational cost for thousands of model runs to bound the topological uncertainty quickly adds up to petabytes of information, which is presently not computationally feasible. Therefore, many researchers have turned to graph theory and ML to develop reduced-order models to represent these systems \cite{srinivasan_quantifying_2018}.

Deep learning (DL)  \cite{lecun_deep_2015} is a special class of machine learning methods that have shown to be very successful in solving many problems in big, spatially patterned and/or time-series data. They currently provide the best solutions to many problems in image recognition \cite{krizhevsky2012imagenet} or natural language processing. Many of these fundamental successes are also successfully applied in other scientific fields such as physics, chemistry, and neuroscience. To accelerate ML development and applications in porous and fractured media, the community needs to establish reliable “ground truth" benchmark datasets because supervised deep learning methods require a large amount of validated data to train models; and the capabilities of the trained classifiers must be assessed quantitatively. In a typical deep learning process, researchers feed image data through neural networks (NNs) such as convolutional NNs (CNNs) and apply optimization algorithms to update the network properties so that the CNN ``learns" to recognize particular patterns. The main application area for developing such methods is processing torrents of photo and video information uploaded on various internet platforms where benchmark datasets are readily available \cite{imgNet, berkelyNet}. Recent applications have successfully adopted these tools for the spatial identification of phases from porous media images \cite{andrew2018quantified, wu2019machine,buckman2017quantifying, fend2021reconstruction, yu2020identification, alqahtani2018deep, alqahtani2020machine,wang2018porous}. In addition, recent advances in generation models such as generative adversarial networks (GANs) show promising results of image generation that can account for multipoint statistics of porous and channel media \cite{kim2021fast} and of very challenging drainage network generated based on network connectivity information \cite{kim2021connectivity}. Morphological geometry and properties of porous and fractured media (e.g., images at pore scale), model parameter values at reservoir scale or known field information such as geological features are essentially tied to spatio-temporal nature of coupled processes of interest, which can be a structured or unstructured data type and spatially and/or temporally correlated. In other words, relevant simulation or experimental data can be seen as images and other physical quantities. An ultimate outcome is that ML methods can help integrate and analyze data faster and accurately with minimized human intervention, which can enable us to build a robust and reliable ML system for accelerating computationally expensive simulations and extracting and interpreting the important features that control the model behavior.

\subsection*{State of the art}
\addcontentsline{toc}{subsection}{State of the art}
Currently, there are many well established numerical methods that can obtain flow properties via three-dimensional pore-scale images or other detailed geometry models: finite volume method, smoothed particle hydrodynamics, finite element method, and the lattice Boltzmann method \cite{meakin_modeling_2009, yoon2015lattice}, and the same is, of course, valid for reservoir-scale simulation \cite{chen_computational_2006}. All simulations in porous media is hindered by the exponential rise in computational demands (time, memory, and CPU/GPU power) with increasing model complexity. When experimental results are required for either simulation input or model validation, we are faced with the issue that experimental measurements are typically sparse and data-fitting has limited validity. Those problems could be potentially remedied by using ML methods that learn from existing data and simulation and perform a robust extrapolation. A number of recent efforts have applied ML to porous media (references follow below). Here, we review some of these attempts with the understanding that this is a fast-moving research field. 

On the smaller scales, a number of recent publications demonstrated the opportunity to apply deep learning models for the physical prediction of either velocity fields (and thus permeability) or permeability directly based on images. The majority has thus far focused on relatively homogeneous media such as sand or sandstones  \cite{kamrava_linking_2019, santos_poreflow-net_2020}, with one method tackling heterogeneous media such as fractured or vuggy \cite{santos_computationally_2021}. Either way, once trained, the methods show an incredible speed up. However, training can be lengthy and requires a lot of data. There is also a lot of trial and error in successful training, and it is not clear how to consistently achieve good results based on (potentially sparse) data only. A recent preprint \cite{zhou2021neural} shows that the majority of the error in the artificial neural networks prediction comes from a failure to observe physical constraints. Even a basic strategy to impose a physical constraint will improve the prediction. 

Some combination of the classical way to impose physical constraints using PDEs and ML is then possibly a winning combination. The energy approach to the solution of PDE via ML is examined in \cite{samaniego2020energy}. This approach enables us to explore different mathematical models by defining the corresponding energies and straightforwardly implementing them. Further, authors in \cite{innes2019differentiable} show how to integrate machine learning tools with traditional physical modeling tools (such as ODE solvers) in a seamless fashion. Authors in \cite{kadeethum2021framework} show how to solve PDEs with highly heterogeneous parameter fields in both the forward and inverse sense using an image-to-image ML framework. The method can also handle discontinuities in the inputs and outputs. Moreover, in \cite{kim2021fast, kim2020efficient} authors developed efficient projection-based nonlinear manifold reduced-order models to reduce computational time. Their work used neural networks for reduced solution representation and incorporated the existing numerical schemes, such as finite element, finite volume, and finite difference methods, to accelerate simulations without losing accuracy. 

Fractured porous media add another level of complexity. Recent advances in high-performance computing have opened the door for flow and transport simulations in large explicit three-dimensional discrete fracture networks \cite{berrone2013pde,erhel2009flow,mustapha2007new,pichot2010mixed,pichot2012generalized,hyman2015dfnworks,hyman2015influence,joyce2014multiscale}. This increase in model fidelity comes at a huge computational cost because of the large number of mesh elements required to represent thousands of fractures (with sizes that can range from mm to km) and properly resolve pressure at fracture intersections where gradients are highest. A combination of graph theory and ML methods have been used to either identify the backbone \cite{hyman2017predictions}, or the small fraction of fractures participating in the physics of flow and transport \cite{hyman2017passage, hyman2018identifying}. These methods use classification or thresholding techniques to identify flow and topological properties that are characteristic of fractures that comprise the backbone for flow. More advanced methods use the graph representation of the fracture network to resolve the flow and transport instead of solving the governing equations on the discretized mesh. These methods have resulted in speedups of up to 4 orders of magnitude, enabling an uncertainty quantification framework for predictive modeling and risk-based decision making. 

For the modeling of reactive transport processes, in addition to the challenge of bridging scales, there are some unique requirements. First, the speciation reactions are repeated at every time step and in each grid cell and are typically most time-consuming. Second, reaction kinetics, especially for minerals, have strong nonlinear dependence on fluid chemistry and local flow and transport features. The third is the need to account for the resulting change in pore structure and feedback on reactive transport processes. Recent advances in ML have been increasingly adopted in reactive transport modeling to address some of the challenges above. Three types of applications have been noted here. First, using ML algorithms to develop surrogate models to replace reactive transport modeling. This approach has been successfully applied to model reactive mixing \cite{MudunuruKarra2021} and electrokinetic remediation of contaminated groundwater \cite{SprocatiRolle2021}. Second, using ML algorithms to replace the most time-consuming steps of reactive transport models, i.e., geochemical calculations \cite{prasianakis2020neural}. Leal et al. \cite{Leal2020} have integrated on-demand ML algorithms for geochemical speciation, using calculations at earlier time steps in the same simulation as training data. Third, using ML algorithms to derive parameters (e.g., reaction rate) for upscaling \cite{prasianakis2020neural}. Finally, a sophisticated example of how ML can be used in reactive transport modeling in an upscaling application is described in  \cite{prasianakis2020neural}.  

\subsection*{Knowledge gaps and challenges}
\addcontentsline{toc}{subsection}{Knowledge gaps and challenges}

\subsubsection*{Knowledge gaps}
\addcontentsline{toc}{subsubsection}{Knowledge gaps}
Key challenges in predictive modeling of flow and transport in fractured and porous media include quantification of the structural and compositional details of the media, identification of relevant physical and chemical processes and how they couple together, and the resulting temporal evolution of media. Advances in high-resolution imaging and high-fidelity models have helped bridge these gaps. Still, the time cost associated with image processing and simulation based on images requires high-performance computing (HPC). It thus can be prohibitively high for the wide adoption of these methods. ML algorithms are expected to be able to accelerate the progress by, e.g., providing faster and more effective image processing and surrogate models. However, for ML algorithms to be an integral component of the workflow for investigating fractured and porous media, we need to build a scientific environment that integrates data, simulation, and prediction efficiently.

We organize the knowledge gaps in the following key groups: (1) data sharing (experimental or synthetic), (2) connecting simulation to data and researchers, (3) property databases that use simulation and data to organize useful information. Those are stepping stones to (4) bridging spatial and temporal scales required to understand porous and fractured media.

\paragraph{\underline{Data Sharing}} 
One of the key current knowledge gaps is the efficient exchange of data between researchers that produce them and those who use them as input or verification in simulation. Further, data often requires processing and clean-up (curation) before it can be reused in traditional simulation or any type of ML. Scientific and engineering opportunities for data-based upscaling can grow exponentially if the methods are combined with access to curated data and scalable computing capabilities required to run simulations and ML applications efficiently. This is why we spend disproportionately more space expanding on the problem of data accessibility and its exchange in research.
 
Storage of experimental datasets has been typically the responsibility of investigators, and traditionally the data has not been readily accessible to other researchers. Some more focused storage is typically associated with specific experimental equipment. For example, imaging centers such as the University of Texas High Resolution Computed Tomography (UTCT) have dedicated data storage. Such centers, however, are under no obligation to keep the data indefinitely, nor can they share it with anyone other than their owners. A study shows that 20 years after a paper is published, $80\%$ of the related data are no longer available due to the researchers moving or to technology transition \cite{vines_availability_2014}. Specific repositories are therefore needed for continued access to curated data. 
 
Online data curation is a fast-moving field, and registries of research data repositories \cite{re3data} as well as search and citation services exist \cite{datacite}. It goes without saying that reproducible research requires open data. However, open data in itself does not mean that researchers training ML algorithms can find and efficiently access a large number of datasets related to a specific problem. There are many sites where you can share any data (typically) as a compressed archive: Mendeley Data \cite{mendeleydata}, Figshare \cite{figshare}, Zenodo \cite{zenodo} and Dryad \cite{dryad}. However, none of those websites prescribed a data format or (compressed) archive organization, and only Dryad performs a quality check (curation). With no consistent formatting, any algorithm's direct usage of the data requires lots of manual work and thus does not scale. Further, when posting to general data-sharing websites, it might be difficult to find all data that is in a specific research field and form a research community around them. Guidelines such as the FAIR (Findability, Accessibility, Interoperability, and Reuse) principles provide a useful framework to develop data repositories for the reuse of scholarly data \cite{wilkinson_fair_2016}.
 
In porous media, there are few dedicated data repositories. For instance, Energy Data Exchange, EDX, \cite{edx} is focused on any data type related to (mostly subsurface) energy. Available data through EDX covers different sources and types, and while users post archives, they are also invited to describe them and enter keywords to promote discoverability. No curation or review of data is done presently. However, EDX is searchable, and each dataset gets a digital object identifier (DOI), thus enabling referencing when reused. Another example is Digital Rocks Portal \cite{prodanovic_digital_2015} with a more narrow focus on imaged porous materials irrespective of imaging modality. Data is curated (reviewed) before posting and issued a DOI, and the community is further engaged by issuing newsletters and organizing mini-courses and contests. Society of Petroleum Engineers has very recently organized an industry data website \cite{spedata} with, at the time of writing this, one dataset. This reflects the fact that the traditional oil and gas industry does not readily share the data. The emergence of such a shared website marks an attempt to break away from that tradition.

\paragraph{\underline{Connecting Simulation to Data and Researchers}}
One of the key knowledge gaps in porous and fractured media simulation is often missing (benchmark) data required for their validation and input parameters. This is why we identified the data as the first knowledge gap and devoted considerable space to elaborating related issues. Secondary issues are consistent simulation code maintenance, their discoverability by their potential users, and their ease of use by a novice user.

Open-source codes are generally available and shared, though they are not all consistently maintained. Codes are also, in essence, text (storage size requirements are not an issue as is the case for data), and sharing the code in itself is straightforward using established version-control repositories such as GitHub \cite{github}. Another efficient option to share codes is to package them into containers, using containerization platforms such as, Docker \cite{merkel2014docker} or CoreOS Rkt \cite{noauthor_CoreOS}. A container is a standalone unit of prepackaged code and all its dependencies, enabling the application to run efficiently and independently of the computing environment. The basic advantage of using containers is that they are portable, lightweight, and secure. Portable - can be used across platforms, lightweight - containers share the machine's OS system kernel and therefore do not require an OS per application, and secure - code and dependencies are isolated in the container and can not contaminate the host OS. All of these practices are common in computer science and software engineering; however, we note that a general scientist or beginning domain researcher is typically not aware of them.

In the porous and fractured media community, Open Porous Media Initiative \cite{opm} notably includes code, documentation, tutorials as well as a selection of field datasets for many (mostly reservoir simulation) tools that encompass everything from meshing to transport simulation. What is missing is providing the same codes compiled and ready to be used on HPC systems and a direct link to available data. Developing that type of cyberinfrastructure is expensive but would expand access to the latest technology and serve as a jumpstart for new researchers. We further elaborate on this idea in the Recommendation section.   

\paragraph{\underline{Property Databases}} 
Spatial and temporal measurements of porous and fractured materials often have to be processed and integrated to obtain useful properties of porous or fractured media. As previously explained, processing of large datasets in porous and fractured media and their integration is non-trivial and requires HPC, and thus we distinguish property databases -  where such processed information is categorized for further use by the research community - from data portals. 

If simulation can easily access data portals, then it is possible to build a property database that can be mined both for experimental data measurements that can serve as inputs to simulators (for instance, permeability lab measurements), or simulated measurements (for instance, permeability could be computed by a simulator based on data in Digital Rocks Portal and then entered into a database). We are currently not aware of such integrated public databases for porous and fractured media that are opened and integrate data from multiple research groups. 

\paragraph{\underline{Bridging Spatial and Temporal Scales}} 
Last but not least, we need novel approaches to successfully bridge the spatial and temporal scales without the loss of scale-specific information. This used to be achieved through multiscale modeling approaches and can be enriched by new machine learning approaches. A hybrid traditional numerical modeling/ML approach is needed where different ML concepts are integrated with well-established numerical methods, and emerging model discovery methods from spatiotemporal systems  \cite{rackauckas2020universal, kaheman2020sindy} have yet to be applied to multiscale and multiphysics processes in porous and fractured media. However, well-organized/accessible data-set libraries in these fields are exceptionally challenging.

How to tackle time-dependent multi-scale problems with strong coupling between scales is an important knowledge gap. These problems arise in porous and fractured media, e.g., in multiphase flows where viscous fingering is driven by pore-scale properties of the medium but impacts larger-scale flow. Another example is the propagation of a fracture in a porous medium. Solutions to these types of problems would have a major benefit on the field.
Multi-scale modeling in heterogeneous porous materials via ML is elaborated in the following section of this report. 

\subsubsection*{Challenges}
\addcontentsline{toc}{subsubsection}{Challenges}
The overarching challenge in porous and fractured media research is integrating individual pieces that are carefully researched but focus on a specific process, on a specific time and length scale. For the most part, the physical intuition of domain scientists comes first in integrating knowledge, and any automated solutions (including ML) come second. Here, we summarize a few challenges that arise from this unique overarching feature and need to be addressed to better integrate ML/artificial intelligence (AI) into porous and fractured media research.

\paragraph{\underline{Data and Research Exchange}} ML needs to learn from many trusted data to capture the physical behavior. We referred to this as needed Data Portals above. However, preparation and labeling of data-sets in geophysics is very challenging as it requires a certain level of knowledge - one needs to be properly trained to label different material phases in a complex morphology (in other words, MSc or Ph.D. training is typically required, or equivalent experience). This means that a standard crowd-sourcing approach is not possible. This becomes even more important as experimental techniques produce extremely large and complex data sets, and there is a serious bottleneck in processing them. Last but not least, hosting and sharing a large volume of data is very expensive - today's datasets need a large amount of storage space and large bandwidth for upload/download and high-performance computing. It is not clear who is supposed to shoulder this investment's expense if everyone is required to post and store data..

\paragraph{\underline{Simulation and ML}} Both simulation and ML in porous and fractured media are computationally intensive. The current hardware possessed by an individual researcher usually cannot handle large problems, especially when dealing with multiple spatial and temporal scales. In ML training, we divide the data into subsets so that the ML model can process the large problem, but this will incur additional errors due to the loss of interactions between different subsets. HPC access is improving in the United States but requires training before incoming students can successfully use it.

A further challenge is a trustworthiness and accuracy. When solving complex engineering problems, one should have a rigorous mathematical model of the underlying physics of the problem or otherwise have a good handle on the accuracy of any numerical solutions. Even if data input for ML is available, we control the inputs to ML training and the design of the NN. The training itself is trial and error (that corrects the inputs or the NN design), even when it produces remarkable results such as provided in the current State-of-the-Art section. We do not fully understand how NN learns, what happens in the hidden layers and why it makes mistakes when applied to new cases (this is an open problem in computer science). First, we need to better measure and define appropriate ML accuracy standards compared to conventional numerical simulations. Second, for coupled processes in porous and fractured media, experimental observations and/or high fidelity models that can be used for training ML models are typically limited to a few specific boundary conditions. When these ML models are applied to different conditions, we are challenged with how we evaluate such models, what is the expected accuracy, and how we incorporate physical solutions.

\paragraph{\underline{Property Databases}} Once data, simulation, and ML approaches are connected and trustworthy, the information and knowledge integrated from both should be collected into porous and fractured media property databases, as elaborated in the Knowledge gaps section. This is a continuous process, and therein lies the challenge of having access to the most up-to-date information.

\paragraph{\underline{Bridging Spatial and Temporal Scales}} The Department of Energy National Laboratories have acted as hubs that provide continuity in developing large-scale simulation tools, including those for porous and fractured media. A notable funding effort was supported by the National Science Foundation that tried to coordinate all of the above aspects in Geosciences via the EarthCube initiative \cite{noauthor_earthcube_nodate, cutcher-gershenfeld_build_2016}, though it is currently winding down. This effort focused on coordinating cyberinfrastructure for data and traditional simulation, and not so much on ML (due to its timing). It is not clear what the future of this effort will be beyond NSF EAR Geoinformatics Program, which is much smaller in scope. EarthCube website \cite{noauthor_earthcube_nodate} lists many efforts across the spectrum of geosciences data (ocean, hydrogeology, fossils, tectonics, earth observation from space) as well as software architecture practices \cite{rubin_recommended_2020}. While there are many good tools developed in the process, they only scratch the surface in the needs of data and research exchange, simulation, or property databases in Geosciences, partly because of vast differences in time/length scales and research tools different Geosciences efforts require.

\subsection*{Recommendations to advance the field in ten years}
\addcontentsline{toc}{subsection}{Recommendations to advance the field in ten years}
\begin{figure}[!h]
    \centering
    \includegraphics[scale = 0.55]{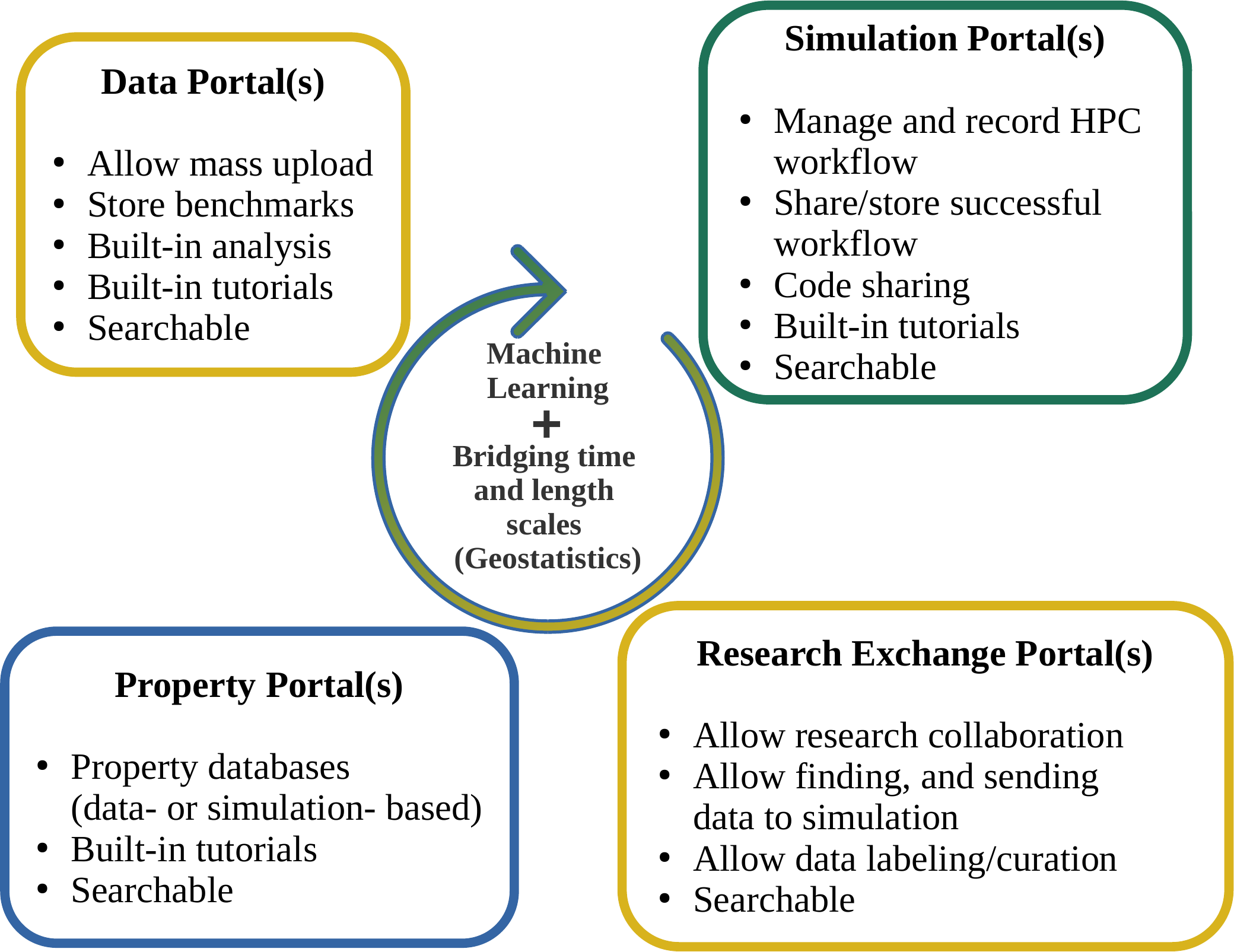}
    \caption{For ML to advance modeling multiscale flow and transport in porous end fractured media or reduce associated computational time, they need to be trained with data. We depict here a desired environment of different portals that are interconnected by the use of machine learning tools, as well as research-community-approved approaches that bridge spatial and temporal scales. These portals represent key knowledge gaps. They provide the platforms for addressing the key challenge of data and research exchange in building multiscale modeling methods for porous and fractured media. Their listed functionality and ML approaches integrating between them represent the recommendations for the advancement of the field. }
    \label{fig: ML-environment}.
\end{figure}

We recognize that in order to address the knowledge gaps and challenges discussed in the previous sections, it is critical to facilitate the collaborations of researchers across different disciplines. Thus \textbf{we recommend the porous and fractured media research community to build an inter-operable cyber-environment of data, simulation, integrated property and research exchange portals} that address knowledge gaps and challenges discussed and enable us to advance ML methods through open-collaboration. See schematic in Figure \ref{fig: ML-environment}. We define a \textbf{portal} to be an online repository with (a friendly) graphical user interface that is accessible to users through a web browser and interoperable (capable of efficient information exchange) with other portals. For each portal in the figure, we list desired functionalities.  \textbf{Machine/deep learning engines are crucial in harvesting information and interacting of all of these elements} thus appearing in the center. Alternatively, ML/DL approaches are considered to be components of all of these portals. The framework illustrated by Figure  \ref{fig: ML-environment} is broadly applicable and can be easily adopted for fields beyond the porous and fractured media community by replacing the mentioning of geostatistics with other relevant terms. Yet, at the same time, the sustainability of each portal is highly dependent on the specific research community (e.g., in porous and fractured media) and its successful communication built around existing knowledge/vocabulary, institutions, or networks.

In the adoption of ML methods in fractured and porous media research, engaging both modelers and experimental researchers will help establish the portals listed in Figure \ref{fig: ML-environment}, and building HPC tools to be used by experimental researchers will broaden the impact of ML methods. It is equally important that integrated collaboration between the fractured and porous media domain scientists and computer scientists would motivate and drive further development of ML methods and lead to better engagement of the broader communities.
%}

To facilitate building the environment suggested in Figure \ref{fig: ML-environment}, we have a number of specific recommendations as follows.
\begin{enumerate}
    \item We recommend co-design of numerical modeling in porous and fractured media and ML methods that help extract parameters and representative equations from available data that bridge spatial and temporal scales. 
    \item We recommend the funding opportunities for collaborative efforts between domain scientists in porous and fractured media and computer science to build data and research exchange portals, simulation portals, and property portals. This ensures that the development of ML/DL modeling methods suitable for porous and fractured media is supported by sufficient high-quality data and domain expertise to bridge spatial and temporal scales. In addition to greater collaborative efforts, there should be more start-up funding grants (with a faster apply-to-get-funded turnaround) that would: a) fund many ideas in a low-risk environment and b) allow for an easy-to-test-and-fail environment. 
    \item Specific attention should be given to the creation of benchmark collections and data labeling so that ML algorithms can thrive. Specific funding should enable the research community to label the datasets that will be further used to characterize porous/ fractured media properties or ML training. Domain expertise is required to curate data (we express this in the form of Research Exchange Portal in Figure \ref{fig: ML-environment}). We need large volumes of labeled data for ML, and despite a few individual attempts to provide a range of datasets for ML training \cite{santos_3d_2021, wang_super_2019}, there is a notable absence of organized benchmark collections in geosciences. Such benchmark collections should be organized and easily used by everyone (including ML codes) and independent of experience level with high-performance computing. 
    \item To better connect the porous and fractured media research communities with unique scientific challenges with tech companies that are major powerhouses for advancing ML/DL techniques, we need to create efficient knowledge transfer mechanisms. One example that can facilitate this transfer, especially in the long run, is to create student internship opportunities in scientific ML in technology companies or domain-specific companies where AI is used and developed. We recognize that a lot of investment in ML is accomplished by companies, and we need more interaction between companies and domain scientists to allow for knowledge transfer.
    \item We recommend creating initiatives that promote scaling teaching of porous and fractured media topics to a broader audience. This is necessary both for the collaborative efforts and success of HPC tools described above and for the community to be more inclusive (see JEDI section)..
\end{enumerate}

In summary, our discussion suggests that tight integration of data, simulation, property databases, and research community, facilitated by machine learning and multidisciplinary training of researchers to manage these resources, is critical to advancing our understanding of porous and fractured media.
 
\section*{Acknowledgments}
\addcontentsline{toc}{subsection}{Acknowledgments}
Sandia National Laboratories is a multi-mission laboratory managed and operated by National Technology and Engineering Solutions of Sandia, LLC., a wholly owned subsidiary of Honeywell International, Inc., for the U.S. Department of Energy’s National Nuclear Security Administration under contract DE-NA0003525. This report, SAND2021-14093 R, describes objective technical results and analysis. Any subjective views or opinions that might be expressed in the paper do not necessarily represent the views of the U.S. Department of Energy or the United States Government.

%references
\addcontentsline{toc}{subsection}{References}
{\renewcommand{\markboth}[2]{}% Remove header adjustment
\printbibliography}
%\printbibliography
\newpage
\end{refsection}
%TA4 - ML in predicting material properties
\begin{refsection}
\begin{center}
\section*{ML in predicting material properties}
\addcontentsline{toc}{section}{ML in predicting material properties}
\author*{Krishna Garikipati$^1$, Lori Graham-Brady$^2$, Vahid Keshavarzzadeh$^3$, Chunhui Li$^4$, Xing Liu$^5$, Piotr Zarzycki$^4$}
\end{center}
\noindent
$^1$ University of Michigan, MI, USA.\\
$^2$ Johns Hopkins University, MD, USA.\\
$^3$ University of Utah, UT, USA.\\
$^4$ Lawrence Berkeley National Laboratory, CA, USA.\\
$^5$ Brown University, RI, USA.\\
\vspace{0.5cm}

\subsection*{Abstract}% (should be max 200 words.)}
\addcontentsline{toc}{subsection}{Abstract}
Machine learning has the potential to accelerate materials discovery by
predicting material properties and leading the tailored design of new materials at reduced computational cost compared to traditional {\it in silico} physics-based simulation methods. However, our ability to develop algorithmic approaches to material design is hampered by the limitations of existing AI approaches in reaching the accuracy expected from the material property predictors.
A new holistic approach is needed to accurately predict material properties across spatial and temporal scales. Specifically, advances are needed to address challenges of ML interpretability, imposition of physicochemical constraints and incorporation of domain knowledge, cost of data generation via theory and experiment, and to meet the expectation that ML models can extrapolate towards unexplored parameter/property space. 

\subsection*{Introduction}
\addcontentsline{toc}{subsection}{Introduction}
There are a number of applications and approaches in which ML is currently employed to predict material properties. In particular,
Deep Neural Networks (DNNs) are widely used in constitutive modelling. This includes representation via Bayesian NNs \cite{zhang2021bayesian}, incorporation of information from lower scale computations such as density functional theory and molecular dynamics \cite{teichert2019machine,teichert2020scale,teichert2021lixcoo2}, direct application of experimental data into DNN, Gaussian Process representations of constitutive models, digital generation of virtual microstructure \cite{bhaduri2021ML} and finally homogenization of material response 
\cite{zhang2020machine}. Significant challenges remain for the problem of material property prediction, in particular those of interpretability and availability of properly curated and managed data.

\subsection*{State of the art}
\addcontentsline{toc}{subsection}{State of the art}
As in many other areas in science and engineering, interpretability of the ML representations presents a major challenge to identifying the physical drivers and mechanisms that most influence material properties \cite{molnar2019}. Some possible approaches that might address these challenges include physics-informed custom descriptors paired with feature weighting attention maps, embedding of invariants, feature selection and encoding, Shapley values\cite{moosavi2020understanding}, self-discovering physical and chemical constraints as patterns-rules that are learned by DNN trained on physically-chemically well constrained dataset\cite{li2021deep}, and physical interpretation of inputs in ML models for mechanical homogenization\cite{hashemi2019novel}. While all of these approaches provide promising directions for the field, there is no doubt that more generalized solutions would be very welcome, and perhaps this is something that expertise in the computer science community could address. 
    
The lack of a single definitive source for standardized, curated and properly documented datasets remains an enormous challenge for the materials community. There are numerous datasets based on different sources, including DFT datasets (MP, A-FLOW,OQMD, etc), experimental datasets (Citrination, MDF, CSD), benchmark datasets \cite{clement2020benchmark, henderson2021benchmark, dunn2020benchmarking}, Github repos from individual articles, and data from journals like {\it Scientific Data} or {\it Data in Brief}. 
Beyond these, there are many web resources, including: pymatgen, The Materials Project, The NIST Materials Data Repository, Materials Cloud and CRAEDL. These resources provide a significant body of highly valuable data to the research community; however, the sheer number of databases leaves the materials researcher with the onerous task of gathering and compiling data from multiple sources. Standardization of these resources and a streamlined interface that can mine data from all of them would be a game-changing capability for materials science.

Inverse approaches to determining material properties have gained in power and use, with significant implications for materials design. These include system inference to determine model forms \cite{wang2019variational, wang2021variational, wang2021inference} and gradient or non-gradient based inverse solution frameworks \cite{wang2021inverse}.

Also of importance are graph-based methods e.g. for fracture, but also for representation, analysis and reduced-order modeling \cite{banerjee2019graph}. Approaches to represent materials information are crucial to move the field forward. Some approaches include: composition-based feature vector versus structure-based feature vectors, extensions to composites as opposed to single phase materials. Other representations that go beyond local influences are also of importance.
 
Material synthesizability, representation of long-range interactions in materials within ML models and modeling across chemical compound spaces represent yet another class of challenges. Across all aspects of ML in materials properties prediction, the community must learn from experimental data. Limiting research to computed data alone reduces the application to only improving efficiency issues. Effective data curation needs to be understood as a challenge across computational and experimental data. There are numerous intriguing applications that can immediately benefit from combining physics-based modeling and machine learning; and designing polycrystaline microstructures is only one of them. This includes extending methods for chemistry where the use of fragments, fingerprints, SMILES etc. approaches is common now to create new molecules unlike inorganic materials where periodicity is a concern, and extending robust representations like SELFIES has interesting potentials for crystal discovery.

\subsection*{Knowledge gaps and challenges}
\begin{figure}[!h]
    \centering
    \includegraphics[scale = 0.60]{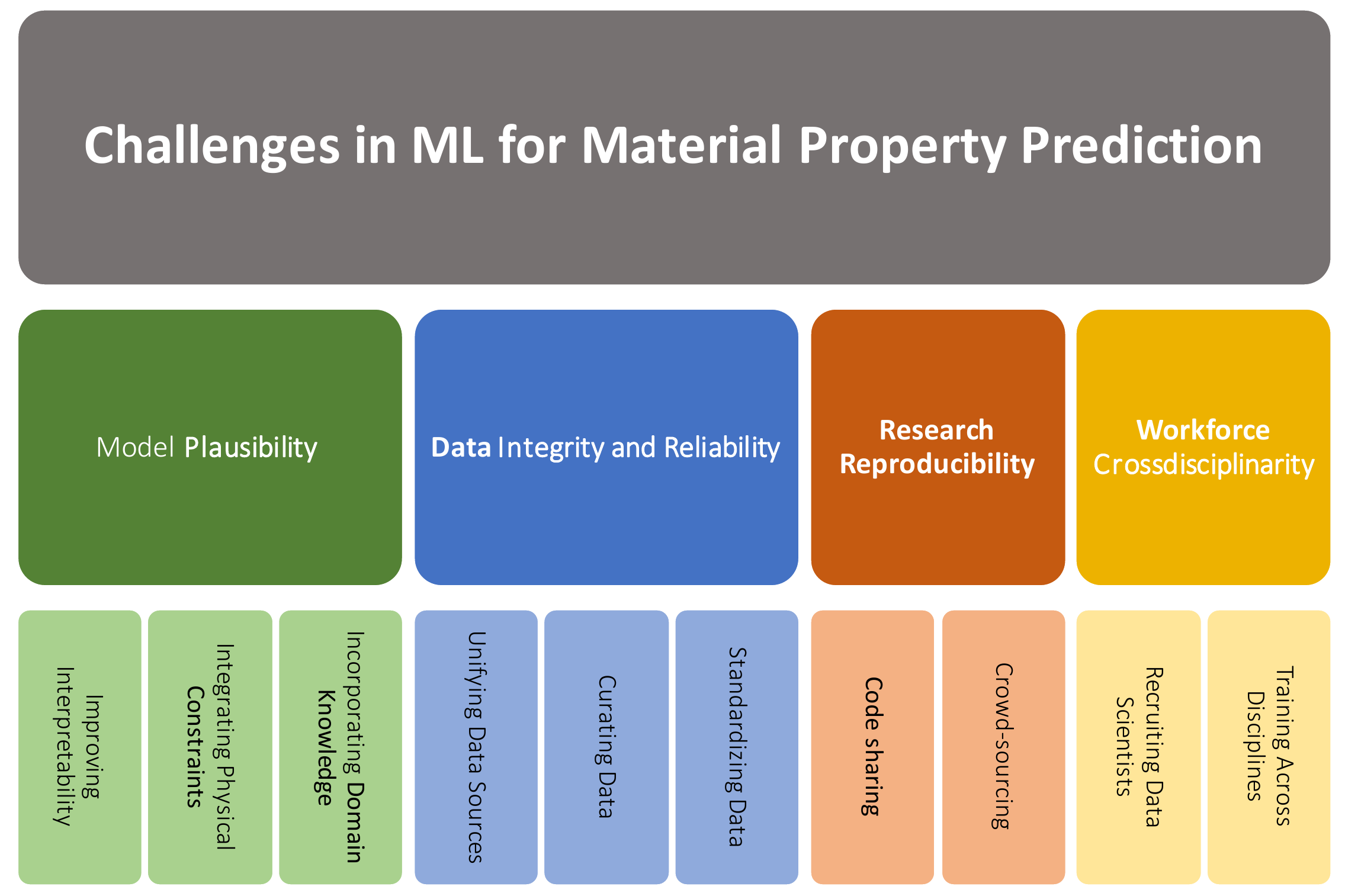}
    \caption{Knowledge gap and challenges in machine learning for material property prediction.}
    \label{fig:TA4}
\end{figure}

\addcontentsline{toc}{subsection}{Knowledge gaps and challenges}
Machine learning methods have become an integral part of our daily lives as a result of advancements in the field. A few impressive applications include speech and image recognition, routine recommender systems, or language translation. In addition, ML has a unique ability to discover patterns in scientific data that can lead to significant scientific advances and potential breakthrough discoveries. However, there are several challenges that must be addressed regarding the limitation of the current ML algorithms and available scientific datasets. In the following we discuss four main challenges including technical and non-technical issues we are facing as a community: 1) how to come up with plausible machine learning models?, 2) how/where to acquire reliable data?, 3) how to verify the integrity of research results? and 4) how to ensure continued development of the junior workforce? These challenges are not limited to our specific topic, but also apply to the previously mentioned topics, as summarized in Figure~\ref{fig:TA4}. 

\paragraph{\underline{Plausible ML Models}}
As stated in Figure \ref{fig:TA4}, there are three main challenges to come up with ML models in scientific fields: incorporating domain knowledge, integrating physical constraints, and improving interpretability. Specifically, porous materials are often anisotropic, with responses and behavior correlated to local and global symmetries. However, current ML algorithms cannot exploit material symmetry. There is a need to develop algorithmic procedures to include short-range order in amorphous phases and long-range order of crystalline lattices. A further challenge is that many materials are not homogeneous across all spatial- and temporal scales relevant for their macroscopic behavior. For example, we need to account for the porosity and permeability of fluid-saturated rocks, the slow transformation of the diagenetically evolving sediments including compaction and fracturing, or multiple possible geochemical reactions spanning a wide range of time scales. 

It remains a challenge to reasonably represent the nano- and microstructure of materials, underlying connectivity or symmetry, and known relationships between these features and physicochemical material properties.

There is also a significant challenge in representing sufficient length and time scale to enable representation of rare events. Many material properties related to failure rely on rare or stochastic events, including mineral dissolution/growth in a saturated environment, a bond breaking at the tip of a slowly growing crack, a microstructural configuration making the region amenable to fatigue crack initiation and growth, or radionuclide decay. Thus, a new generation of ML algorithms needs to become sensitive to the infrequent events that are critically important for overall material responses.  

Finally, the current ML algorithms have intrinsic difficulty with extrapolation beyond the parameter space explored during the training phase and solving inverse problems with more than one unique solution. Thus ML practitioners are faced with the dilemma of choosing an optimal algorithm for a given problem, selecting a minimum complexity of the model, and customizing existing algorithms to account for the challenges discussed above. Approaches to quantify the uncertainties associated with ML models provide some promise for tackling this issue, but such tools are far from mature \cite{olivier2021}.

\paragraph{\underline{Data Integrity and Reliability}}
There are several challenges related to the scientific data and to exploring them using ML. In Figure \ref{fig:TA4}, we divided the challenges into three categories: 1) data standardization, 2) data curation, and 3) data unification. Many of the ML limitations discussed above also apply to existing scientific datasets. For example, the scale separation, representation of the physicochemical structure of materials and their time-dependent evolution should be addressed either at the data preparation step, constructing ML algorithms, or to a different degree in both phases. 
Another major challenge is the limited amount of publicly/academically available data and little knowledge transfer between data creators and data scientists. Data scarcity, questionable reliability, and non-standardized formats also hamper data utilization. With the rapidly increasing ability to measure and store massive amounts of data, the scientific community will benefit from having a dedicated data storage facility with robust protocols for data representation, quality and version controls, and augmentation procedures.  

\paragraph{\underline{Research Reproducibility}}
One significant challenge in any research whether computational or experimental is the replication of process and results. As illustrated in Figure \ref{fig:TA4}, crowd-sourcing and code sharing have been important challenges in reproducing scientific discoveries. With the advent of code sharing platforms such as Github, these issues have been circumvented to some extent; however, there is still no systematic way of testing and analyzing codes and research developments. This challenge might be exacerbated in fields that inherently necessitate fast development and communication such as computer science, and may carry over into machine learning for science.

\paragraph{\underline{Workforce Crossdisciplinarity}}
The final challenge of predicting material property is workforce crossdisciplinarity (see Figure \ref{fig:TA4}). In academia, it is a challenge to maintain workforce continuity as postdocs and students join and leave frequently. It is imperative to establish a system which keeps track of code developments and documents for new students to learn projects and build  skill-sets easily and quickly. Another significant challenge for academic research is the unbalanced reward system in the job market. Individuals with expertise in data sciences are in high demand in many fields, including industry centered around data sciences for more highly resourced industries like the financial and information technology sectors. The lower salaries in academia compared to these industries has led large numbers of skilled graduate students and postdoctoral trainees to pursue careers outside the academic world. 

\subsection*{Recommendations to advance the field in ten years}
\addcontentsline{toc}{subsection}{Recommendations to advance the field in ten years}
There have been significant developments in machine learning in various domains that have had a significant, and largely, positive impact on many aspects of society. We are optimistic about the potential of machine learning in advancing our knowledge of fundamental domains including materials science. 

One example of potential advancement is the enabling of autonomous materials design/ discovery where the decisions (such as finding the optimized route to the target) is made in an AI-driven fashion. There are other similar examples such as developing workflows for autonomous discovery comprised of small (manageable) individual building blocks, 
developing game theoretic approaches which combine experiment, theory and computation, ways to explore high dimensional design space (e.g. in material discovery) more effectively or similarly acquiring optimal experimental design in a sequential manner. 

In the same vein as above i.e. the optimal experimental design, ML can potentially help determining informative experiments such as identifying the important scales and/or underlying mechanisms, and provide guidance for complex problems. An example of autonomous experimental design, i.e. generating experimental data in real time is Electronic Lab Notebooks (ELNs). Similarly, ML can be effective in imposing smart constraints on our search toward definitive objectives such as realizable, synthesizable and manufacturable materials systems. Such efforts on data-driven discoveries can also be relevant to assimilating uncertain data to make a better performing models. 

Another important subject which could be implemented more effectively via and within ML pipelines is crowd-sourcing. Crowd-sourcing is a highly effective approach for performing heavy tasks such as material discovery. Crowd-sourcing reduces human bias, improves the quality and interpretation of data and enables analyzing large datasets. 

There are also other recommendations such as building a platform/community for sharing data, exploiting the potential of Generative Adversarial Networks (GANs) for data generation that could potentially replace costly scientific simulations, and utilization of ML/AI for literature review and summarizing the findings/state of the science in the literature. 

In summary, there are numerous important areas of basic science and applied research where combining physics-based modeling and machine learning can significantly advance our understanding of existing porous materials and promote the design of new, lightweight, or smart materials that harness the advantages of porous nano and micro structures.

%references
\addcontentsline{toc}{subsection}{References}
{\renewcommand{\markboth}[2]{}% Remove header adjustment
\printbibliography}
\newpage
\end{refsection}
%JEDI
\begin{refsection}
\hypersetup{
    colorlinks=true,
    linkcolor=blue,
    filecolor=magenta,      
    urlcolor=blue,
    citecolor=blue,
   % pdftitle={Overleaf Example},
    pdfpagemode=FullScreen,
    }

\section*{Recommendations to promote JEDI (Justice, Equity, Diversity, and Inclusion)}
\addcontentsline{toc}{section}{Recommendations to promote JEDI (Justice, Equity, Diversity, and\\ Inclusion)}
The general idea of Justice, Equity, Diversity, and Inclusion (JEDI) has received considerable attention throughout past decades as these notions can play a significant role in advancing science and general human knowledge leading to the betterment of the world and society as a whole.

It is agreed among most people that a diverse group addressing a problem is always more successful. This is especially true in the context of ML for heterogeneous materials, where a wide range of knowledge and perspectives across fields is needed to understand the underlying physics and to implement efficient ML models. Although the participation of underrepresented communities and women in STEM, and retention of young researchers in academia and research labs is constantly improving, we still have a lot of work ahead of us to reach \textit{justice}, \textit{equity}, \textit{diversity}, and \textit{inclusion} in our scientific communities. 

\subsection*{Improving the Existing Educational Systems}
\addcontentsline{toc}{subsection}{Improving the Existing Educational Systems}
We have found that the ML communities that we participate in are strongly lacking in all forms of diversity. Less than 14\% of AI papers on arXiv were written by women \cite{stathoulopoulos2019gender}, and a 2018 report estimated that just 12\% of AI researchers are women \cite{chin2018ai}. Lack of participation from underrepresented communities can be traced back to inadequacy in K-12 education. This indicates students from underrepresented communities do not have an equal footing when they start to apply for colleges and universities, and they get screened out from the beginning. Furthermore, due to the steep learning curve and fast-paced nature of higher education, the students who managed to get into the universities find themselves left behind and failing to reach the next level. One size fits all solution for these problems is hard to envision, and it will probably not work. To overcome these issues, grass-root initiatives and systematic advancements to our educational systems are needed. In the improved system, children from all backgrounds, social, and economic classes are taught high-quality lessons from kindergarten to higher education.

\subsection*{Retention of Minorities and Women}
\addcontentsline{toc}{subsection}{Retention of Minorities and Women}
In the context of ML, with its strong applicability to industry, it is especially hard to retain early career, mid-career, and senior scientists in academia and national labs. The combination of benefits, salary, and lack of control over location can make careers in academic settings less appealing than industry positions. These problems particularly confound the issues of recruiting and retaining a diverse staff and/or faculty. 

We further recognize that the challenge also arises from the leaks along the training pipeline. In addition to outreach and internship opportunities for under-represented community students, it is imperative to ensure a positive experience and the success of under-represented students that we already have, through supportive mentoring programs, etc. In particular, there is a lack of permanent positions in academia and the research community. Many are not able to sustain their research due to a lack of secure positions and funding. Lack of permanent jobs and funding, in turn, leads to a smaller pool of Ph.D. students. The brightest students opt for industry jobs as they offer higher-paid and more secure jobs.

We also note that female faculty members are 33\% more likely to have full-time working spouses than male faculty members leading to an extensive increase of pressure on female faculty members  \cite{jacobs2004overworked}. Such pressure can make higher salaries and benefits of industrial careers more appealing.

One recommendation is to pay specific attention to benefits that can aid early career researchers with families, such as childcare support provided by some agencies (i.e., NIH grants \cite{nih}). Childcare support can increase the accessibility of workshops, conferences, and meetings. 

Many postdocs have reported being discouraged from taking, or not having access, to maternity leave. Specifically, one study found 44\% of externally funded postdocs have no access to leave for birth parents \cite{ledford2017us}. The issue is compounded for those in historically underrepresented groups, who are more likely to be discouraged from taking a parental leave \cite{ledford2017us}. When parental leave is available, the decrease in academic output due to leave can increase the pressure to move to a career viewed more compatible with having a family \cite{ysseldyk2019leak}.

This pressure has been compounded by the care-taking responsibilities necessitated by the COVID-19 pandemic, reducing the number of publications by female authors in 2021 compared to 2019 \cite{ribarovska2021gender, viglione2020women, gayet2021female}. This reduction in publishing can and will have long-term impacts on the careers of female researchers. 

Simultaneously, the COVID-19 pandemic has increased the prevalence of flexible work environments, remote work, and virtual and hybrid collaborations and conferences. In particular, virtual conferences have increased the participation of students and women significantly \cite{skiles2020beyond}. Those who have family obligations or have limited ability to travel due to disabilities or funding can now participate in virtual conferences while in-person conferences may be inaccessible. We encourage the field to examine the role of virtual and hybrid work and events in increasing inclusivity within the field. As the field begins to come out of the COVID-19 pandemic and adopt new modes of work, we also encourage consideration of the lessons learned from the pandemic to be continued as we move forward.

\subsection*{Creating Equal Opportunities for All}
\addcontentsline{toc}{subsection}{Creating Equal Opportunities for All}
Equal accessibility is another key item for promoting JEDI in STEM. The development of the field of ML is accompanied by the availability of open-source algorithms, which is desirable for JEDI and should be adopted in other fields as well. However, it should also be noted that accessibility to computational time and other resources needed to learn and run ML algorithms are not to be taken for granted. It is therefore critical to develop mechanisms that allow access to necessary resources at different stages of related learning and research. At the highest level,  we need to reduce the level of elitism in academia, such as famous researchers and their collaborators have a much easier path (less scrutiny) to publishing. We thus specifically recommend for the research community and funding agencies to:
\begin{enumerate}
    \item Support open source and FAIR data for easy access to everyone.
    \item Support outreach to under-represented groups and training that enable the use of existing research resources.
    \item Provide easier access to smaller, start-up grants (i.e. shorter grant application and turnaround time) for all would enable faster turnaround of the ideas, easier change of research area and retraining, and quicker failure of the ideas that do not work. 
    \item Promote collaborative spaces. Specifically, we recommend finding expert collaborators from different fields that can leverage their experience and resources to establish collaborative spaces. Such collaborative hubs will be beneficial to students and young researchers alike, opening new avenues for research, collaboration, and knowledge exchange. This can also be done through so-called ``startup research funding'', which will be used to organize short-term collaborative spaces for postdocs and young researchers to work on a unique problem.   
\end{enumerate}

\subsection*{Convergence of Academia, Industry, and National Labs}
\addcontentsline{toc}{subsection}{Convergence of Academia, Industry, and National Labs}
We are glad to see the rise of variate open-source platforms across communities, e.g., CCSNet~\cite{wen2021ccsnet}, Aflow.org, Materials Commons~\cite{puchala2016materials} etc., which enable researchers from different backgrounds to explore machine learning tools. Meanwhile, researchers should be aware of the different learning curves for ML and domain knowledge. The entry-level ML toolbox for a researcher with a background in computational mechanics and sciences is usually low, thanks to the user-friendly toolkit and online tutorials. However, it is not trivial for a computer scientist to learn numerical schemes, conservation laws, multiscale modeling, and other domain knowledge. As we realized the challenges and opportunities in this field, it is important to integrate physics knowledge with machine learning skills. People in industry, national labs, and academia should be integrated at different levels. Although the goals from industry, National Labs, and academia are quite different, industries and national labs can provide important perspectives, calibration, and funding opportunities for academia. It is important to identify the state-of-the-art and benchmark problems. One recommendation is to define a few use cases that the community can discuss in terms of accuracy, convergence, and computational costs. A good idea will be to provide a well-defined problem with a reference solution to a community and let people solve it with rewards sponsored by National Labs and/or industry. A good example is the Sandia Fracture Challenge \cite{Boyce2014, Boyce2016, Kramer2019, Spear2019TheTS}, which has systematically brought various groups together and provided educational tools for students and researchers across the globe. 

To tackle the challenges mentioned above, different academic communities need to work together. The theoretic front of neural network (NN) approximation needs to be pushed by mathematicians. We hope mathematical-based theory can help design the NN training like some recent work from the numerical analysis framework (e.g., ~\cite{siegel2020approximation,weinan2020towards}). Meanwhile, experimentalists can help us understand what is happening when ML does not work and see how physics and modeling interact. ML, as well as finite volume/element numerical methods, should be added to the toolbox of computational engineering and sciences. Different academic communities, as well as industries and National Labs, should join forces towards defining the benchmark problems, sharing widely not only about what works but also about what does not work and rewarding the participants in challenge competitions, which are inclusively open to all people across the globe.

\subsection*{Democratization of ML in Various Scientific and Engineering Communities}
\addcontentsline{toc}{subsection}{Democratization of ML in Various Scientific and Engineering Communities}
Within the ML community for material science, there have already existed various interdisciplinary/collaborative efforts among computer scientists and material scientists. Such efforts can be augmented by extending materials research (or more specifically material informatics) to geographical regions in which there are fewer Research-based universities. Similarly, there are primarily undergraduate institutions (PUI) that are active in research and could benefit from JEDI initiatives within scientific and engineering communities. Learning resources in material informatics such as \href{https://molssi.org/}{https://molssi.org/} are effective to expand the reach of computational materials science research and as a result to promote JEDI. The popular ML platforms such as Tensorflow \cite{tensorflow2015}, Pytorch \cite{NEURIPS20199015}, etc. have also significantly contributed to the democratization of ML research in various fields including materials science. Taking advantage of these publicly available platforms, many junior-level researchers (including undergraduate students) can now effectively integrate their disciplinary research with state-of-the-art ML developments. 

\subsection*{Organizing Diverse Forums/Workshops/Conferences}
\addcontentsline{toc}{subsection}{Organizing Diverse Forums/Workshops/Conferences}
In addition to the above existing efforts, organizing forums such as the AmeriMech workshop that led to this report is an effective approach to
bringing researchers from different backgrounds and stages together. Ensuring the diversity of workshop panels is also crucial for promoting JEDI. Moreover, students and early-career researchers also benefit from small workshops and small group discussions like those that take place during ICERM and SIAM conferences. Another example is creating a list of minorities or women in computational fields (e.g., \href{https://awmadvance.org/research-networks/wincomptop-women-in-computational-topology/people/}{Women in Computational Topology}) from where keynote speakers for various workshops and conferences can be selected.

%references
{\renewcommand{\markboth}[2]{}% Remove header adjustment
\printbibliography}
\addcontentsline{toc}{subsection}{References}
%\printbibliography
\newpage
\end{refsection}
%Acknowledgement
\begin{refsection}
\section*{Acknowledgements}
\addcontentsline{toc}{section}{Acknowledgements}
The organizing committee would like to express sincere appreciation to the National Academies of Sciences, Engineering and Medicine and the U.S. National Committee on Theoretical and Applied Mechanics for sponsoring our AmeriMech symposium on ``Machine learning in heterogeneous porous materials". We would also like to thank Prof. Tarek Zohdi, the chairs, scribes, and participants. Special thanks to the authors of the report as well as the reviewers: Prof. Ellen Kuhl from Stanford University, Prof. Miguel Bessa from TU Delf University, Tarek I. Zohdi from UC Berkeley, and Dr. Youngsoo Choi from Lawrence Livermore National Laboratory whose feedbacks have significantly improved the quality of our report. Without our chairs, scribes, authors, and reviewers, this report would not have been possible.

We would also like to thank the Department of Mechanical Engineering at The University of Utah for assistance with the logistics and behind the scene preparations for the workshop. Special thanks to Drs. Bozo Vazic and Christopher Creveling for all their supports before, during, and after the workshop.

\textbf{\href{https://newell.mech.utah.edu}{Pania Newell}}
The University of Utah\\
\textbf{\href{https://www.brown.edu/research/projects/crunch/george-karniadakis}{George Karniadakis}}
Brown University\\\textbf{\href{https://www.lanl.gov/search-capabilities/profiles/hari-viswanathan.shtml}{Hari Viswanathan}}
Los Alamos National Laboratory
\end{refsection}
%Appendix
 \begin{refsection}
\newpage
\vspace*{\fill}
\begingroup
\begin{center}
\section*{Appendix}
\addcontentsline{toc}{section}{Appendix}
\begin{itemize}
\centering
\Large
    \item[A.] Symposium Agenda
    \item[B.] List of Participants
\end{itemize}
\end{center}
\endgroup
\vspace*{\fill}
\newpage
%\addcontentsline{toc}{subsection}{Symposium Agenda}
\includepdf[pages=-]{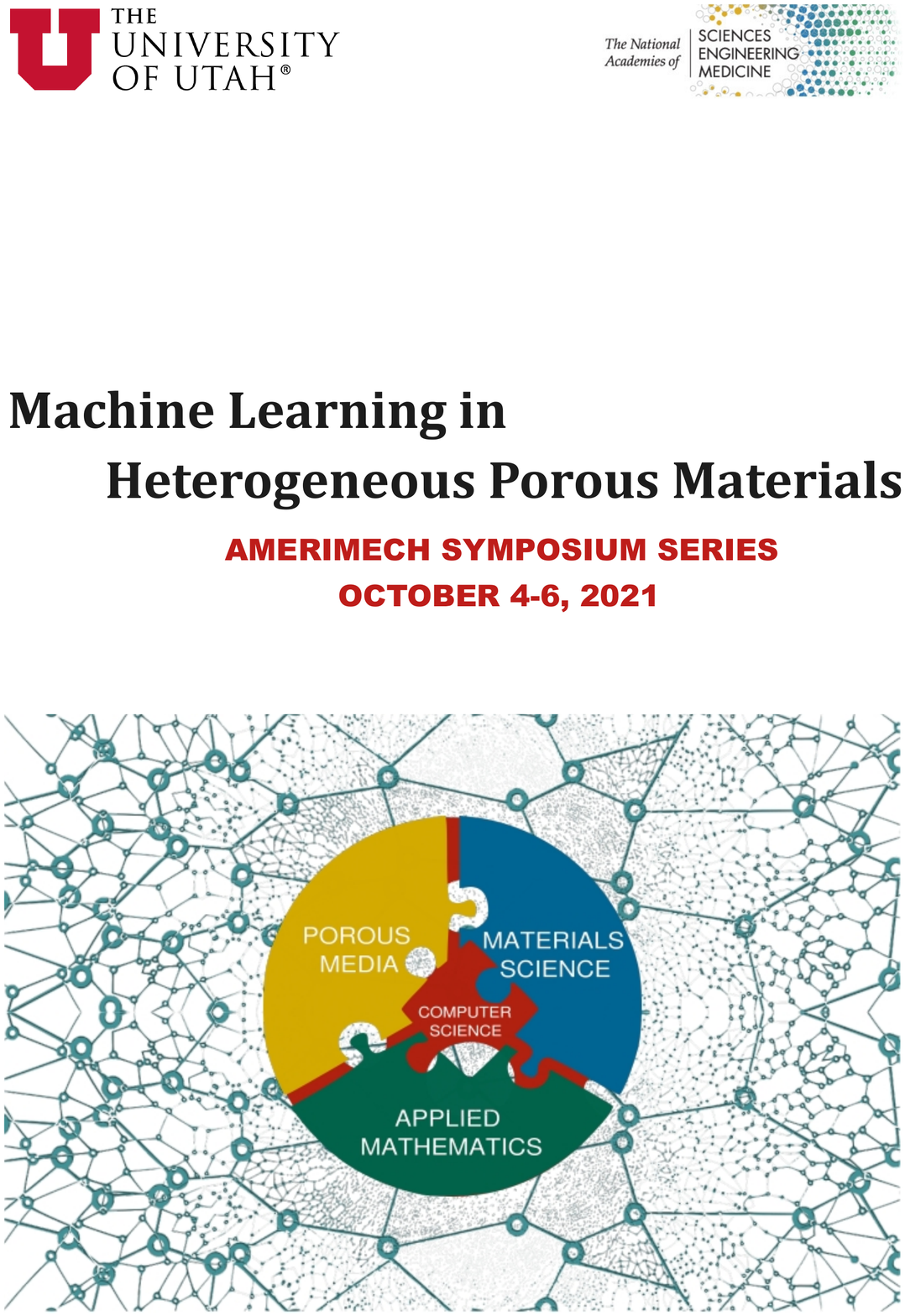}
\end{refsection}
%Participants
\begin{refsection}
\setcounter{page}{1}
\fancyfoot[C]{B-\thepage} 
\begin{center}
\section*{List of Participants}
%\addcontentsline{toc}{subsection}{List of Participants}
\end{center}
\normalsize \begin{longtable}{ p{.35\textwidth}  p{.65\textwidth} }
\textbf{Name/Surname}	&	\textbf{Affiliation} \vspace{0.1in}	\\
Abdullah Cihan	&	Lawrence Berkeley National Lab	\\
Abigail Hunter	&	Los Alamos National Laboratory	\\
Alexandre M Tartakovsky	&	University Of Illinois Urbana-Champaign	\\
Amanda Howard	&	Pacific Northwest National Laboratory	\\
Andrew Lew	&	Massachusetts Institute of Technology	\\
Andrew M. Stuart	&	California Institute of Technology	\\
Azadeh sheidaei	&	Iowa State University	\\
Bashir Khoda	&	The University of Maine	\\
Bei Wang	&	The University of Utah	\\
Beijun Shen	&	Johns Hopkins University	\\
Bikash Chandra Dey	&	The University of Utah	\\
Bo Ni	&	Massachusetts Institute of Technology	\\
Cedric G Fraces	&	Stanford university	\\
Chunhui Li	&	Lawrence Berkeley National Laboratory	\\
Claire Olivia	&	N/A	\\
Danai Koutra	&	University of Michigan	\\
Daniel M Tartakovsky & Stanford University	\\
Daniel O'Malley	&	Los Alamos National Laboratory	\\
Dongwoon Shin	&	Max Planck Institute for Polymer Research	\\
Ehsan Motevali Haghighi	& McMaster University	\\
Farhin Tabassum	&	Bangladesh University of Engineering and Technology (BUET)	\\
Gao Huajian 	&	Brown University	\\
Gowri Srinivasan	&	Los Alamos National Lab	\\
Grace Gu	&	University of California, Berkeley	\\
Hae Young Noh	&	Stanford University	\\
Hamid Ghasemi	&	Howard University	\\
Hang Deng	&	Lawrence Berkeley National Laboratory	\\
Hannah Lu	&	Stanford university	\\
Henry Boateng	&	San Francisco State University	\\
Hongkyu Yoon	&	Sandia National Laboratories	\\
Ilenia Battiato	&	Stanford university	\\
Ilias Bilionis 	&	Purdue University	\\
Jin Yang	&	University of Wisconsin-Madison	\\
Kai Jin	&	Massachusetts Institute of Technology, The University of Utah 	\\
Lei Wang	&	University of the District of Columbia	\\
Lori Graham-Brady	&	Johns Hopkins University	\\
Lucy Zhang	&	Rensselaer Polytechnic Institute	\\
Maarten de Hoop	&	Rice University	\\
Markus J Buehler	&	Massachusetts Institute of Technology	\\
Marta D Elia	&	Sandia National Laboratories	\\
Maryam Aliakbari	&	Northerm Arizona University	\\
Michael Penwarden	&	The University of Utah	\\
Mike Kirby	&	The University of Utah	\\
Ming Dao	&	Massachusetts Institute of Technology	\\
Mohamed Mehana	&	Los Alamos National Lab	\\
Mohammad Hashemi	&	Iowa State University	\\
Mohammadreza Soltany Sadrabadi	&	Northern Arizona University	\\
Nathan Kutz	&	University of Washington	\\
Nikolaos Bouklas	&	Cornell University	\\
Peyman Mostaghimi	&	University of New South Wales - Australia	\\
Piotr Zarzycki	&	Lawrence Berkeley National Laboratory	\\
Pramod Bhuvankar	&	Lawrence Berkeley National Laboratory	\\
Robert Michael Ii Kirby	&	University of Utah	\\
Rohit Batra	&	Argonne National Lab	\\
Sangryun Lee	&	University of California, Berkeley	\\
Seyed Jamaleddin Mostafavi Yazdi	&	Research Scientist	\\
Seyed Mohamad Moosavi	&	FU-Berlin/EPFL	\\
Somnath Ghosh	&	Johns Hopkins University	\\
Song Zhigong	&	Institute of High Performance Computing (IHPC)	\\
Surya Kalidindi	&	Georgia Tech	\\
Susu Xu	&	Stony Brook University	\\
Taylor Sparks	&	The University of Utah	\\
Vahid Keshavarzzadeh	&	The University of Utah	\\
Vipin kumar	&	University of Minnesota	\\
WaiChing Sun	&	Columbia University	\\
Wei Lu	&	Massachusetts Institute of Technology	\\
Xiangyu Sun	&	University of Wisconsin Madison	\\
Xing Liu	&	Brown University	\\
Xingjie Li	&	University of North Carolina Charlotte	\\
Yu-Chuan Hsu	&	Massachusetts Institute of Technology	\\
Yue Yu	&	Lehigh University	\\
Zeqing Jin	&	University of California, Berkeley	\\
Zhenze Yang	&	Massachusetts Institute of Technology	\\
Zhizhou Zhang	&	University of California, Berkeley	\\
\end{longtable}
\end{refsection}
\end{document}